\documentclass{article}
\usepackage[dvipsnames]{xcolor}
\usepackage{iclr2021_conference,times}

\usepackage{amsmath,amsfonts,bm}

\def\eqref#1{equation~\ref{#1}}

\def\1{\bm{1}}

\DeclareMathAlphabet{\mathsfit}{\encodingdefault}{\sfdefault}{m}{sl}
\SetMathAlphabet{\mathsfit}{bold}{\encodingdefault}{\sfdefault}{bx}{n}

\usepackage[utf8]{inputenc} %
\usepackage[T1]{fontenc}    %
\usepackage{hyperref}       %
\usepackage{url}            %
\usepackage{booktabs}       %
\usepackage{amsfonts}       %
\usepackage{nicefrac}       %
\usepackage{microtype}      %
\usepackage{amsmath}
\usepackage{standalone}
\usepackage{wrapfig}
\usepackage{adjustbox}
\usepackage[linesnumbered,ruled,vlined]{algorithm2e}

\SetCommentSty{mycommfont}
\usepackage{graphicx}
\usepackage{hyperref}
\usepackage{enumitem}
\usepackage{mwe}

\usepackage{placeins}
\usepackage{floatrow}

\usepackage{subcaption}
\usepackage{tikz}
\usetikzlibrary{positioning}
\usetikzlibrary{arrows}
\usetikzlibrary{shapes.geometric}
\usetikzlibrary{shapes.multipart}
\usetikzlibrary{trees}
\usetikzlibrary{calc}
\renewcommand{\vec}[1]{\bm{\mathbf{#1}}}

\newcommand\blfootnote[1]{%
  \begingroup
  \renewcommand\thefootnote{}\footnote{#1}%
  \addtocounter{footnote}{-1}%
  \endgroup
}

\title{Predicting Infectiousness \\ for Proactive Contact Tracing}

\author{\textbf{Yoshua Bengio}\textsuperscript{*\textit{af}} $\;$ \textbf{Prateek Gupta}\textsuperscript{*\textit{abh}} $\;$ \textbf{Tegan Maharaj}\textsuperscript{*\textit{ag}} $\;$ \textbf{Nasim Rahaman}\textsuperscript{*\textit{ac}} $\;$ \textbf{Martin Weiss}\textsuperscript{*\textit{ag}} \\
\textbf{Tristan Deleu}\textsuperscript{$\dagger$\textit{af}} $\;$ \textbf{Eilif Muller}\textsuperscript{$\dagger$\textit{af}} $\;$ \textbf{Meng Qu}\textsuperscript{$\dagger$\textit{ai}} $\;$ \textbf{Victor Schmidt}\textsuperscript{$\dagger$\textit{a}} $\;$ \textbf{Pierre-Luc St-Charles}\textsuperscript{$\dagger$\textit{a}} \\
\textbf{Hannah Alsdurf}\textsuperscript{$\ddagger$\textit{d}} $\;$ \textbf{Olexa Bilanuik}\textsuperscript{$\ddagger$\textit{a}} $\;$ \textbf{David Buckeridge}\textsuperscript{$\ddagger$\textit{e}} $\;$ \textbf{G\'aetan Marceau Caron}\textsuperscript{$\ddagger$\textit{a}} \\
\textbf{Pierre-Luc Carrier}\textsuperscript{$\ddagger$\textit{a}} $\;$ \textbf{Joumana Ghosn}\textsuperscript{$\ddagger$\textit{a}} $\;$ \textbf{Satya Ortiz-Gagne}\textsuperscript{$\ddagger$\textit{a}} $\;$ \textbf{Chris Pal}\textsuperscript{$\ddagger$\textit{a}} $\;$ \textbf{Irina Rish}\textsuperscript{$\ddagger$\textit{af}} \\
\textbf{Bernhard Sch\"olkopf}\textsuperscript{$\ddagger$\textit{c}} $\;$ \textbf{Abhinav Sharma}\textsuperscript{$\ddagger$\textit{e}} $\;$ \textbf{Jian Tang}\textsuperscript{$\ddagger$\textit{ai}} $\;$ \textbf{Andrew Williams}\textsuperscript{$\ddagger$\textit{a}}
}

\iclrfinalcopy
\begin{document}
\maketitle
\begin{abstract}

\blfootnote{*,$\dagger$,$\ddagger$ Equal contributions, alphabetically sorted; \textsuperscript{a}Mila, Qu\'ebec; \textsuperscript{b}University of Oxford; \textsuperscript{c}Max-Planck Institute for Intelligent Systems T\"ubingen; \textsuperscript{d}University of Ottawa; \textsuperscript{e} McGill University; \textsuperscript{f}Universit\'e de Montreal; \textsuperscript{g}\'Ecole Polytechnique de Montreal; \textsuperscript{h}The Alan Turing Institute; \textsuperscript{i}HEC Montr\'eal}
The COVID-19 pandemic has spread rapidly worldwide, overwhelming manual contact tracing in many countries and resulting in widespread lockdowns for emergency containment.  Large-scale \textbf{digital contact tracing (DCT)}\footnote{All \textbf{bolded} terms are defined in the Glossary; Appendix 1.} has emerged as a potential solution to resume economic and social activity while minimizing  spread of the virus. Various DCT methods have been proposed, each making trade-offs between privacy, mobility restrictions, and public health.
The most common approach, \textbf{binary contact tracing (BCT)}, models infection as a binary event, informed only by an individual's test results, with corresponding binary recommendations that either all or none of the individual's contacts quarantine. 
BCT ignores the inherent uncertainty in contacts and the infection process, which could be used to tailor messaging to high-risk individuals, and prompt proactive testing or earlier warnings. It also does not make use of observations such as symptoms or pre-existing medical conditions, which could be used to make more accurate infectiousness predictions.  %
In this paper, we use a recently-proposed COVID-19 epidemiological simulator to develop and test methods that can be deployed to a smartphone to locally and proactively predict an individual's infectiousness (risk of infecting others) based on their contact history and other information, while respecting strong privacy constraints. Predictions are used to provide personalized recommendations to the individual via an app, as well as to send anonymized messages to the individual's contacts, who use this information to better predict their own infectiousness, an approach we call \textbf{proactive contact tracing (PCT)}. 
Similarly to other works, we find that compared to no tracing, all DCT methods tested are able to reduce spread of the disease and thus save lives, even at low adoption rates, strongly supporting a role for DCT methods in managing the pandemic. Further, we find a deep-learning based PCT method which improves over BCT for equivalent average mobility, suggesting PCT could help in safe re-opening and second-wave prevention.\footnote{Open source code available at \url{https://github.com/mila-iqia/COVI-ML}} %
\end{abstract}

\section{Introduction} %
\label{sec:introduction}
Until pharmaceutical interventions such as a vaccine become available, control of the COVID-19 pandemic relies on nonpharmaceutical interventions such as lockdown and social distancing. While these have often been successful in limiting spread of the disease in the short term, these restrictive measures have important negative social, mental health, and economic impacts.  \textbf{Digital contact tracing (DCT)}, a technique to track the spread of the virus among individuals in a population using smartphones, is an attractive potential solution to help reduce growth in the number of cases and thereby allow  more economic and social activities to resume while keeping the number of cases low.

Most currently deployed DCT solutions use \textbf{binary contact tracing (BCT)}, which sends a quarantine recommendation to all recent contacts of a person after a positive test result. 
While BCT is simple and fast to deploy, and most importantly can help curb spread of the disease \citep{lowadoption}, epidemiological simulations by \citet{hinch2020effective} suggest that using
only one bit of information about the infection status can lead to quarantining many healthy individuals while failing to quarantine infectious individuals. Relying only on positive test results as a trigger is also inefficient for a number of reasons:
 (i) Tests have high false negative rates~\citep{li2020false}; 
 (ii) Tests are administered late, only after symptoms appear, leaving the asymptomatic population, estimated 20\%-30\% of cases
 \citep{gandhi2020asymptomatic}, likely untested; 
 (iii) It is estimated that infectiousness is highest \textit{before} symptoms appear, well before someone would get a test~\citep{heneghan2020sars}, thus allowing them to infect others before being traced, 
 (iv) Results typically take at least 1-2 days, and
 (v) In many places, tests are in limited supply. 
 
Recognizing the issues with test-based tracing, \citet{gupta2020covisim} propose a rule-based system leveraging other input clues potentially available on a smartphone (e.g. symptoms, pre-existing medical conditions), an approach they call \textbf{feature-based contact tracing (FCT)}. %
Probabilistic (non-binary) approaches, using variants of belief propagation in graphical models, e.g. \citep{baker2020probabilistic,satorras2020neural, briers2020risk}, could also make use of features other than test results to improve over BCT, although these approaches rely on knowing the social graph, either centrally or via distributed  exchanges between nodes. The latter solution may require many bits exchanged between nodes (for precise probability distributions), which is
 challenging both in terms of privacy and bandwidth. 
Building on these works, we propose a novel FCT methodology we call \textbf{proactive contact tracing (PCT)}, in which we use the type of features proposed by \cite{gupta2020covisim} as inputs to a predictor trained to output \textit{proactive} (before current-day) estimates of expected infectiousness (i.e. risk of having infected others
in the past and of infecting them in the future). 
The challenges of privacy and bandwidth motivated our particular form of \textbf{distributed inference} where we pretrain the predictor offline and do not assume that the messages exchanged are probability distributions, but instead just informative inputs to the node-level predictor of infectiousness.

\iffalse
Our method is designed to be used on a smartphone, allowing for the use of self-reported symptoms, pre-existing conditions, and anonymized contact histories to predict the individual's risk (expected infectiousness) over the last 14 days\footnote{For contacts more than 14 days ago, sending updates would not generally make any difference: the other person would already have enough evidence that either they did or did not get infected.}. 
Predicted infectiousness is used in two ways: (1) to warn the individual via graded recommendations so they can change their behavior appropriately, 
 and (2) to inform contacts of changes in estimated risk via anonymized messages. These messages about the estimated expected infectiousness of previous contacts allow others to estimate their own infectiousness, and thus change their behavior. Note that the infectiousness that matters for the messages exchanged between users who met $t$ days ago is the infectiousness $t$ days ago, whereas the infectiousness that matters for changing one's behavior now is one's current infectiousness. PCT is thus based on predicting both current and past infectiousness to make it possible to provide early, `proactive' warning signals.
\fi
We use a recently proposed COVID-19 agent-based simulation testbed \citep{gupta2020covisim} called COVIsim to compare PCT to other contact tracing methods under a wide variety of conditions. We develop deep learning predictors for PCT in concert with a professional app-development company, ensuring inference models are appropriate for legacy smartphones. By leveraging the rich individual-level data produced by COVIsim to train predictors offline, we are able to perform individual-level infectiousness predictions locally to the smartphone, with sensitive personal data never required to leave the device. We find deep learning based methods to be consistently able to reduce the spread of the disease more effectively, at lower cost to mobility, and at lower adoption rates than other predictors.  These results suggest that deep learning enabled PCT could be deployed in a smartphone app to help produce a better trade-off between the spread of the virus and the economic cost of mobility constraints than other DCT methods, while enforcing strong privacy constraints.

\subsection{Summary of technical contributions}
\begin{enumerate}[leftmargin=0.8cm]
\item We examine the consequential problem of COVID-19 infectiousness prediction and propose a new method for contact tracing, called proactive contact tracing (see Sec. \ref{sec:probabilistic-contact-tracing}).  %

\item In order to perform distributed inference with deep learning models, we develop an architectural scaffolding whose core is any set-based neural network. 
We embed two recently proposed networks, namely Deep Sets \citep{zaheer2017deep} and Set Transformers \citep{lee2018set} and evaluate the resulting models via the COVIsim testbed \citep{gupta2020covisim} (see Sec. \ref{sec:distributed_inference}). 

\item To our knowledge the combination of techniques in this pipeline is entirely novel, and of potential interest in other settings where privacy, safety, and domain shift are of concern. Our training pipeline consists of training an ML infectiousness predictor on the domain-randomized output of an agent-based epidemiological model, in several loops of retraining to mitigate issues with (i) non-stationarity and (ii) distributional shift due to predictions made by one phone influencing the input for the predictions of other phones. (see Sec. \ref{subsec:training-procedure})

\item To our knowledge this is the first work to apply and benchmark a deep learning approach for probabilistic contact tracing and infectiousness risk assessment. 
We find such models are able to leverage weak signals and patterns in noisy, heterogeneous data to 
 better estimate infectiousness compared to binary contact tracing and rule-based methods (see Sec. \ref{sec:experiments})

\end{enumerate}

\section{Proactive Contact Tracing}
\label{sec:probabilistic-contact-tracing}
\begin{figure}[h]
\centering
\includegraphics[width=\textwidth]{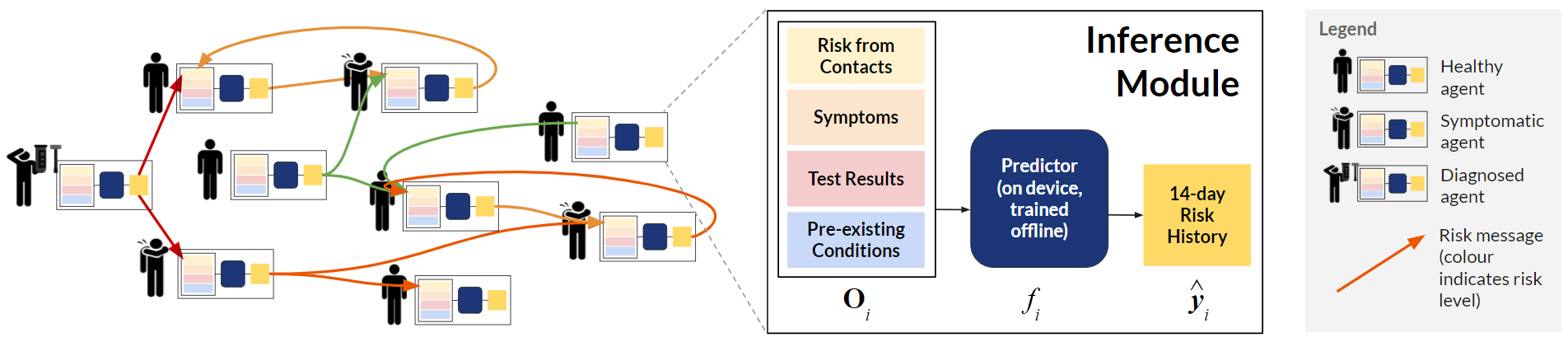}%
\caption{\textbf{Proactive contact tracing overview.} Diagram showing \textbf{Left:} the propagation of anonymized, graded (non-binary) risk messages between users and \textbf{Inset:} overview of the inference module deployed to each user's phone. The inference module for agent $i$ takes in observables $\mathbf{O}_i$, and uses a pretrained predictor $f$ to estimate that agent's risk (expected infectiousness)  for each of the last 14 days. Anonymized selected elements of this risk vector are sent as messages to appropriate contacts, allowing them to proactively update their own estimate of expected infectiousness. } %
\label{fig:PCToverview}
\end{figure}

\textbf{Proactive contact tracing (PCT)} is an approach to digital contact tracing which leverages the rich suite of features potentially available on a smartphone (including information about symptoms, preexisting conditions, age and lifestyle habits if willingly reported) to compute proactive estimates of an individual's expected infectiousness. These estimates are used to (a) provide an individualized recommendation and (b) propagate a graded risk message to other people who have been in contact with that individual (see Fig.~\ref{fig:PCToverview}). This stands in contrast with existing approaches for contact tracing, which are either binary (recommending all-or-nothing quarantine to contacts), or require centralized storage of the contact graph or other transfers of information which are incompatible with privacy constraints in many societies. 
Further, the estimator runs locally on the individual's device, such that any sensitive information volunteered does not need to leave the device. 

In Section~\ref{sec:problem_setup}, we formally define the general problem PCT solves. In Section~\ref{sec:privacy_approach}, we describe how privacy considerations inform and shape the design of the proposed framework and implementation. Finally, in Section~\ref{sec:distributed_inference}, we introduce deep-learning based estimators of expected infectiousness, which we show in Section~\ref{sec:experiments} to outperform DCT baselines by a large margin. 

\subsection{Problem Setup} \label{sec:problem_setup}
We wish to estimate \textbf{infectiousness} $y_i^{d'}$ of an agent $i$ on day $d'$, given access to \emph{locally observable information} $\vec{O}_i^d$ now on day $d\geq d'$ and over the past $d_{max}$ days ($d'\geq d-d_{max}$)
Some of the information available on day $d$ is static, including reported age, sex, pre-existing conditions, and lifestyle habits (e.g. smoking), denoted $g_i$, the
health profile. Other information is measured
each day: the health status $h_i^d$ of
reported symptoms and known test results. Finally, $\vec{O}_i^d$
also includes information about encounters in the last
$d_{max}$ days, grouped in $e_i^d$ for day $d$. Thus:
\begin{align}
    \vec{O}_i^d = (g_i, h_i^d, e_i^d, h_i^{d-1}, e_i^{d-1},
    \ldots h_i^{d-d_{max}}, e_i^{d-d_{max}})
\end{align}
 The information $e_i^{d'}$ about the encounters  from day $d'$ is subject to privacy constraints detailed in Section~\ref{sec:privacy_approach} but provides
 indications about the estimated infectiousness of the persons
 encountered at $d'$, given the last available
 information by these contacts as of day $d$, hence
 these contacts try to estimate their past infectiousness a posteriori.
 Our goal is thus to model the \textit{history} of the agent's infectiousness (in the last $d_{max}$ days), which is what enables the recommendations of PCT to be \emph{proactive} and makes it possible for an infected asymptomatic person to receive a warning from their contact even before they develop symptoms, because
 their contact obtained sufficient evidence that they
 were contagious on the day of their encounter.
 Formally, we wish to model 
$P_{\theta}(\vec{y}_i^{d} | \vec{O}_i^{d})$,
where $\vec{y}_i^d=(y_i^d,y_i^{d-1},\ldots,y_i^{d-d_{max}})$ is the
vector of present and past infectiousness of agent $i$ and 
$\theta$ specifies the parameters of the predictive
model. In our experiments we only estimate conditional
expectations with a predictor $f_\theta$, with $\vec{\hat y}_i^d = (\hat y_i^d, ..., \hat y_i^{d - d_{max}}) = f_\theta(\vec{O}_i^d)$ an estimate of the conditional expected per-day present and past infectiousness $\mathbb{E}_{P_\theta}[\vec{y}_i^d | \vec{O}_i^d]$. 

The predicted expected values are used in two ways. First, they are used to generate messages transmitted on day $d$ to contacts involved in encounters on day $d' \in (d-d_{max},d)$. 
These messages contain the estimates $\hat{y}_i^{d'}$ of the expected infectiousness of $i$ at day $d'$,  quantized to 4 bits for privacy reasons discussed in section~\ref{sec:privacy_approach}.
Second, the prediction for today $\hat y_i^d$ is also used to form 
a discrete recommendation level $\zeta_i^d \in \{0, 1, ..., n\}$ regarding the behavior of agent $i$ at day $d$ via a recommendation mapping $\psi$, i.e. $\zeta_i^d = \psi(\hat y_i^d)$. \footnote{The recommendations for each level are those proposed by \citet{gupta2020covisim}, hand-tuned by behavioural experts to lead to a reduction in the number of contacts. Full compliance with recommendations is not assumed.}  
At $\zeta_i^d = 0$ agent $i$ is not subjected to any restrictions, $\zeta_i^d = 1$ is baseline restrictions of a post-lockdown scenario (as in summer 2020 in many countries), $\zeta_i^d = n$ is full quarantine (also the behaviour recommended for contacts of positively diagnosed agents under BCT), and intermediate levels interpolate between levels 1 and $n$.

Here we make two important observations about contact tracing: (1) There is a trade-off between decelerating spread of disease, measured by the \textbf{reproduction number $R$}, or as number of cases,\footnote{Note that the the number of cases is highly non-stationary; it grows exponentially over time, even where $R$, which is in the exponent, is constant.} and minimizing the degree of restriction on agents, e.g., measured by the average number of contacts between agents.
Managing this tradeoff is a social choice which involves not just epidemiology but also economics, politics, and the particular weight different people and nations may put on individual freedoms, economic productivity and public health. The purpose of PCT is to improve the corresponding \textbf{Pareto frontier}. A solution which performs well on this problem will encode a policy that contains the infection while applying minimal restrictions on healthy individuals, but it may be that different methods are more appropriate depending on where society stands on that tradeoff. 
(2) A significant challenge comes from the feedback loop between agents: observables $\vec{O}_i^d$ of agent $i$ depend on the predicted infectiousness histories and the pattern of contacts generated by the behavior $\zeta_j$ of \textit{other agents} $j$. This is compounded by privacy restrictions that prevent us from knowing which agent sent which message; we discuss our proposed solution in the following section. %

\vspace{-0.25 cm}

\subsection{Privacy-preserving PCT}\label{sec:privacy_approach}
\vspace{-0.25 cm}

One of the primary concerns with the transmission and centralization of contact information is that someone with malicious intent could identify and track people. Even small amounts of personal data could allow someone to infer the identities of individuals by cross-referencing with other sources, a process called \textbf{re-identification} \citep{el2011systematic}.  Minimizing the number of bits being transmitted
and avoiding information that makes it easy to triangulate people is a protection
against \textbf{big brother attacks}, where a central authority with access to the data
can abuse its power, as well as against \textbf{little brother attacks}, where malicious
individuals (e.g. someone you encounter) could use your information against you.
Little brother attacks include \textbf{vigilante attacks}: harassment, violence, hate crimes, or stigmatization against individuals which could occur if infection status was revealed to others. Sadly, there have been a number of such attacks related to COVID-19 \citep{russell2020rise}. To address this, PCT operates with 
(1) no central storage of the contact graph; 
(2) de-identification \citep{10.1142/S0218488502001648} and encryption of all data leaving the phones;
(3) informed and optional consent to share information (only for the purpose of improving the predictor); and 
(4) distributed inference which can achieve accurate predictions without the need for a central authority. 
Current solutions~\citep{DCT-list} share many of these goals, but are restricted to only binary CT. Our solution achieves these goals while providing individual-level graded recommendations.

Each app records \textbf{contacts} which are defined by
\cite{canada-public-health-prolonged-exposure} as being an encounter which lasts at least 15 minutes with a proximity under 2 meters. Once every 6 hours, each application processes contacts, predicts infectiousness for days $d$ to $d_{max}$, and may update its current recommended behavior. If the newly predicted infectiousness history differs from the previous prediction on some day (e.g. typically because the agent enters new symptoms, receives a negative or positive test result, or receives a significantly updated infectiousness estimate), then the app creates and sends small update messages to all relevant contacts. These heavily quantized messages reduce the network bandwidth required for message passing as well as provide additional privacy to the individual. We follow the communication and security protocol for these messages introduced by \citet{whitepaper}. We note that a naive method would tend to \textit{over-estimate} infectiousness in these conditions, because repeated encounters with the same person (a very common situation, for example people living in the same household) should carry a \textit{lower} predicted infectiousness than the same number of encounters with different people. To mitigate this over-estimation while not identifying anyone, messages are clustered based on time of receipt and risk level as in  \citet{gupta2020covisim}.%

\section{Methodology for Infectiousness Estimation}
\label{sec:methodology-for-infectiousness-estimation}
\subsection{Distributed Inference} \label{sec:distributed_inference}
Distributed inference networks seek to estimate marginals by passing messages between nodes of the graph to make
predictions which are consistent with each other (and with the local evidence available at each node). Because messages in our framework are highly constrained by privacy concerns in that we are prevented from passing distributions or continuous values between identifiable nodes, typical distributed inference approaches (e.g. loopy belief propagation \citep{murphy1999loopy}, variational message passing \citep{winn2005variational}, or expectation propagation \citep{Minka2001_Expectation_Propagation}) are not readily applicable. We instead propose an approach to distributed inference which uses a trained ML predictor for estimating these marginals
(or the corresponding expectations). The predictor $f$ is trained offline on simulated data to predict expected infectiousness from the locally available 
information $\vec{O}_i^d$ for agent $i$ on day $d$. The choice of ML predictor is informed by several factors. First and foremost, we expect the input $e_i^d$ (a component of $\mathbf{O}_i^d$) to be variable-sized, because we do not know \emph{a priori} how many contacts an individual will have on any day. Further, privacy constraints dictate that the contacts within the day cannot be temporally ordered, implying that the quantity $e_i^d$ is set-valued. Second, given that we are training the predictor on a large amount of domain-randomized data from many epidemiological scenarios (see Section~\ref{subsec:training-procedure}), we require the architecture to be sufficiently expressive. Finally, we require the predictor to be easily deployable on edge devices like legacy smartphones.%

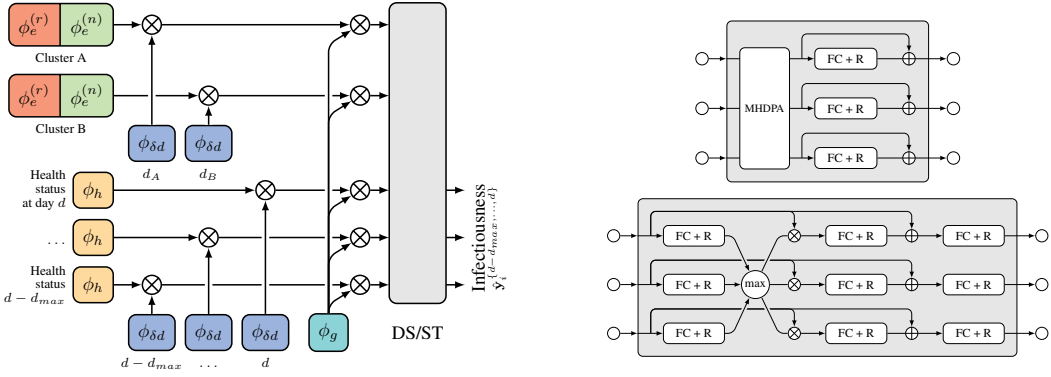
\begin{figure}[t]
\centering
\begin{subfigure}[c]{0.49\textwidth}
  \centering
  \begin{tikzpicture}[remember picture, bigo/.style={inner sep=0pt, fill=white, circle, minimum size=1em, draw=black, thick}, inputs/.style={fill=white, draw=black, rectangle, rounded corners, thick, inner sep=5pt, font=\normalsize}, cluster inputs/.style={inputs, fill=none, rectangle split, rectangle split parts=2, rectangle split horizontal, rectangle split part fill={Red!50, YellowGreen!50}}, health inputs/.style={inputs, fill=Dandelion!50, anchor=east}, day inputs/.style={inputs, fill=NavyBlue!40}, global inputs/.style={inputs, fill=BlueGreen!50}, y=2.5em]
    \node[cluster inputs, label={below:\scriptsize{Cluster A}}] (cluster_A) at (0, 0) {$\phi_{e}^{(r)}$ \nodepart{two} $\phi_{e}^{(n)}$};
    \node[cluster inputs, label={below:\scriptsize{Cluster B}}] (cluster_B) at (0, -1.5) {$\phi_{e}^{(r)}$ \nodepart{two} $\phi_{e}^{(n)}$};

    \coordinate (tmp_health_0) at (0, -3.5 - 0);
    \node[health inputs, label={[align=flush right, font=\scriptsize]left:Health\\[-0.1em]status\\[-0.1em]at day $d$}] (health_0) at (tmp_health_0 -| cluster_A.east) {$\phi_{h}$};
    
    \coordinate (tmp_health_1) at (0, -3.5 - 1);
    \node[health inputs, label={[align=flush right, font=\scriptsize]left:\\[-0.1em]$\ldots$}] (health_1) at (tmp_health_1 -| cluster_A.east) {$\phi_{h}$};
    
    \coordinate (tmp_health_2) at (0, -3.5 - 2);
    \node[health inputs, label={[align=flush right, font=\scriptsize]left:Health\\[-0.1em]status\\[-0.1em]$d - d_{max}$}] (health_2) at (tmp_health_2 -| cluster_A.east) {$\phi_{h}$};

    \coordinate[right=2em of cluster_A] (tmp_otimes_A1);
    \node[bigo] (otimes_A1) at (tmp_otimes_A1) {};
    \draw[-, thick] (otimes_A1.north west) -- (otimes_A1.south east);
    \draw[-, thick] (otimes_A1.south west) -- (otimes_A1.north east);
    
    \coordinate[right=3em of tmp_otimes_A1] (otimes_A2);
    \node[bigo] (otimes_B2) at (cluster_B -| otimes_A2) {};
    \draw[-, thick] (otimes_B2.north west) -- (otimes_B2.south east);
    \draw[-, thick] (otimes_B2.south west) -- (otimes_B2.north east);

    \node[bigo] (otimes_21) at (health_2 -| otimes_A1) {};
    \draw[-, thick] (otimes_21.north west) -- (otimes_21.south east);
    \draw[-, thick] (otimes_21.south west) -- (otimes_21.north east);
    
    \node[bigo] (otimes_12) at (health_1 -| otimes_A2) {};
    \draw[-, thick] (otimes_12.north west) -- (otimes_12.south east);
    \draw[-, thick] (otimes_12.south west) -- (otimes_12.north east);
    
    \coordinate[right=3em of otimes_A2] (otimes_A3);
    \node[bigo] (otimes_03) at (health_0 -| otimes_A3) {};
    \draw[-, thick] (otimes_03.north west) -- (otimes_03.south east);
    \draw[-, thick] (otimes_03.south west) -- (otimes_03.north east);
    
    \draw[-latex, thick] (cluster_A) -- (otimes_A1);
    \draw[-latex, thick] (cluster_B) -- (otimes_B2);
    \draw[-latex, thick] (health_0) -- (otimes_03);
    \draw[-latex, thick] (health_1) -- (otimes_12);
    \draw[-latex, thick] (health_2) -- (otimes_21);
    
    \coordinate[right=5em of otimes_A3] (tmp_otimes_A4);
    \node[bigo] (otimes_A4) at (tmp_otimes_A4) {};
    \draw[-, thick] (otimes_A4.north west) -- (otimes_A4.south east);
    \draw[-, thick] (otimes_A4.south west) -- (otimes_A4.north east);
    
    \node[bigo] (otimes_B4) at (cluster_B -| tmp_otimes_A4) {};
    \draw[-, thick] (otimes_B4.north west) -- (otimes_B4.south east);
    \draw[-, thick] (otimes_B4.south west) -- (otimes_B4.north east);
    
    \foreach \i in {0, ..., 2} {
        \node[bigo] (otimes_\i4) at (health_\i -| tmp_otimes_A4) {};
        \draw[-, thick] (otimes_\i4.north west) -- (otimes_\i4.south east);
        \draw[-, thick] (otimes_\i4.south west) -- (otimes_\i4.north east);
    };
    
    \draw[-latex, thick] (otimes_A1) -- (otimes_A4);
    \draw[-latex, thick] (otimes_B2) -- (otimes_B4);
    \draw[-latex, thick] (otimes_03) -- (otimes_04);
    \draw[-latex, thick] (otimes_12) -- (otimes_14);
    \draw[-latex, thick] (otimes_21) -- (otimes_24);
    
    \node[day inputs, below=1em of otimes_B2, label={below:\scriptsize{$d_{B}$}}] (cluster_day_B) {$\phi_{\delta d}$};
    \node[day inputs, label={below:\scriptsize{$d_{A}$}}] (cluster_day_A) at (cluster_day_B -| otimes_A1) {$\phi_{\delta d}$};

    \node[day inputs, below=1em of otimes_21, label={below:\scriptsize{$d - d_{max}$}}] (health_day_2) {$\phi_{\delta d}$};
    \node[day inputs, label={below:\scriptsize{$d$}}] (health_day_0) at (otimes_03 |- health_day_2) {$\phi_{\delta d}$};
    \node[day inputs, label={[label distance=4pt]below:\scriptsize{$\ldots$}}] (health_day_1) at (otimes_12 |- health_day_2) {$\phi_{\delta d}$};

    \coordinate[right=1em of otimes_A4] (tmp1_dsst);
    \coordinate[right=3em of tmp1_dsst] (tmp2_dsst);
    \coordinate (dsst_tl) at (cluster_A.north -| tmp1_dsst);
    \coordinate (dsst_br) at (health_2.south -| tmp2_dsst);
    \draw[draw=black, fill=gray!20, rounded corners, thick] (dsst_tl) rectangle (dsst_br);
    
    \draw[-latex, thick] (otimes_A4) -- (otimes_A4 -| dsst_tl);
    \draw[-latex, thick] (otimes_B4) -- (otimes_B4 -| dsst_tl);
    \foreach \i in {0, ..., 2}
        \draw[-latex, thick] (otimes_\i4) -- (otimes_\i4 -| dsst_tl);

    \node[global inputs, right=1em of health_day_0] (global) {$\phi_{g}$};

    \foreach \i in {A, B, 0, 1, 2}
        \coordinate[below=0.5em of otimes_\i4] (tmp_global_\i);
    
    \draw[-, line width=4pt, draw=white] (global) -- (global |- tmp_global_A) -- (otimes_A4);
    
    \foreach \i in {A, B, 0, 1, 2}
        \draw[-latex, thick, rounded corners] (global) -- (global |- tmp_global_\i) -- (otimes_\i4);

    \draw[-, line width=4pt, draw=white] (cluster_day_A) -- (otimes_A1);
    \draw[-latex, thick] (cluster_day_A) -- (otimes_A1);

    \draw[-, line width=4pt, draw=white] (cluster_day_B) -- (otimes_B2);
    \draw[-latex, thick] (cluster_day_B) -- (otimes_B2);
    
    \draw[-, line width=4pt, draw=white] (health_day_2) -- (otimes_21);
    \draw[-latex, thick] (health_day_2) -- (otimes_21);
    
    \draw[-, line width=4pt, draw=white] (health_day_1) -- (otimes_12);
    \draw[-latex, thick] (health_day_1) -- (otimes_12);
    
    \draw[-, line width=4pt, draw=white] (health_day_0) -- (otimes_03);
    \draw[-latex, thick] (health_day_0) -- (otimes_03);
    
    \draw[-latex, thick] (otimes_04 -| dsst_br) -- ++(0:1em);
    \draw[-latex, thick] (otimes_14 -| dsst_br) -- ++(0:1em) node[pos=1, font=\normalsize, anchor=north, align=flush center, rotate=90] {Infectiousness \\[-0.2em]\scriptsize{$\hat{\mathbf{y}}_{i}^{\{d - d_{max}, \ldots, d\}}$}};
    \draw[-latex, thick] (otimes_24 -| dsst_br) -- ++(0:1em);

    \coordinate (dsst) at ($(dsst_tl)!0.5!(dsst_br)$);
    \node[] at (global -| dsst) {DS/ST};
\end{tikzpicture}
  \label{fig:model-architecture-1}
\end{subfigure}\hfill
\begin{subfigure}[c]{0.42\textwidth}
  \centering
  \begin{tikzpicture}[remember picture, bigo/.style={inner sep=0pt, fill=white, circle, minimum size=1em, draw=black, thick}]
    \def\imgnumrows{2}

    \node[rectangle, draw=black, thick, fill=gray!20, rounded corners, minimum height=1em, minimum width=5em, inner sep=1em] (background) at (0, 0) {\begin{tikzpicture}[fc/.style={rectangle, draw=black, fill=white, thick, inner sep=5pt, font=\normalsize}, y=4em, remember picture]
        \foreach \i in {0, ..., \imgnumrows} {
            \coordinate (l_\i) at (0, \i);
        }
        
        \foreach \i in {0, ..., \imgnumrows} {
            \coordinate[right=5em of l_\i] (m_\i);
            \coordinate[above=2em of m_\i] (ma_\i);
            \node[fc, right=1em of m_\i] (fc_\i) {FC + R};
            \draw[-latex, thick] (m_\i) -- (fc_\i);
            
            \node[bigo, right=2em of fc_\i] (oplus_\i) {};
            \draw[-, thick] (oplus_\i.north) -- (oplus_\i.south);
            \draw[-, thick] (oplus_\i.west) -- (oplus_\i.east);
            
            \draw[-, thick] (fc_\i) -- (oplus_\i);
            \draw[-latex, thick] (m_\i) -- (ma_\i) -- (ma_\i -| oplus_\i) -- (oplus_\i);
        }

        \coordinate (MHDPA_tl) at (l_0 |- fc_\imgnumrows.north);
        \coordinate[right=4em of MHDPA_tl] (MHDPA_tr);
        \coordinate (MHDPA_br) at (MHDPA_tr |- fc_0.south);
        
        \draw[fill=white, thick] (MHDPA_tl) rectangle (MHDPA_br) node[pos=0.5, inner sep=0pt, minimum size=0pt, font=\small] {MHDPA};
        
        \foreach \i in {0, ..., \imgnumrows}
            \draw[-, thick] (MHDPA_br |- m_\i) -- (m_\i);
    \end{tikzpicture}};

    \foreach \i in {0, ..., \imgnumrows} {
        \node[bigo, left=2.5em of l_\i] (input_\i) {};
        \draw[-latex, thick] (input_\i) -- (background.west |- l_\i);
        \draw[-, thick] (background.west |- l_\i) -- (l_\i);

        \node[bigo, right=2.5em of oplus_\i] (output_\i) {};
        \draw[-latex, thick] (oplus_\i) -- (output_\i);
    };
\end{tikzpicture}\\[5pt]
  \begin{tikzpicture}[remember picture, bigo/.style={inner sep=0pt, fill=white, circle, minimum size=1em, draw=black, thick}]
    \def\imgnumrows{2}

    \node[rectangle, draw=black, thick, fill=gray!20, rounded corners, minimum height=1em, minimum width=5em, inner sep=1em] (background) at (0, 0) {\begin{tikzpicture}[fc/.style={rectangle, draw=black, fill=white, thick, inner sep=5pt, font=\normalsize}, y=4em]
        \foreach \i in {0, ..., \imgnumrows} {
            \coordinate (l_\i) at (0, \i);
            \node[fc, right=1em of l_\i] (fcl_\i) {FC + R};
        }
        
        \coordinate[right=2.5em of fcl_0] (fcl_0_right);
        \coordinate (fcl_n_right) at (fcl_\imgnumrows -| fcl_0_right);
        \node[circle, thick, draw=none, fill=none, inner sep=2pt, minimum size=0pt] (max) at ($(fcl_0_right)!0.5!(fcl_n_right)$) {\phantom{max}};
        \coordinate[right=2em of max] (max_right);

        \foreach \i in {0, ..., \imgnumrows} {
            \node[bigo] (otimes_\i) at (fcl_\i -| max_right) {};
            \draw[-, thick] (otimes_\i.north east) -- (otimes_\i.south west);
            \draw[-, thick] (otimes_\i.north west) -- (otimes_\i.south east);

            \coordinate (tmp_midl_\i) at (fcl_\i -| max.west);
            \coordinate (midl_\i) at ($(fcl_\i.east)!0.5!(tmp_midl_\i)$);
            
            \coordinate (tmp_midr_\i) at (fcl_\i -| max.east);
            \coordinate (midr_\i) at ($(tmp_midr_\i)!0.5!(otimes_\i.west)$);

            \node[fc, right=2em of otimes_\i] (fcm_\i) {FC + R};
            \draw[-latex, thick] (otimes_\i) -- (fcm_\i);
        }

        \coordinate[right=2em of fcm_0] (fcm_right);
        \foreach \i in {0, ..., \imgnumrows} {
            \node[bigo] (oplus_\i) at (fcl_\i -| fcm_right) {};
            \draw[-, thick] (oplus_\i.north) -- (oplus_\i.south);
            \draw[-, thick] (oplus_\i.west) -- (oplus_\i.east);

            \draw[-, thick] (fcm_\i) -- (oplus_\i);
            \node[fc, right=2em of oplus_\i] (fcr_\i) {FC + R};
            \draw[-latex, thick] (oplus_\i) -- (fcr_\i);
            
            \draw[-latex, thick] (l_\i) -- (fcl_\i);
            \coordinate[above=2em of l_\i] (la_\i);
            \draw[-latex, thick] (l_\i) -- (la_\i) -| (otimes_\i);
            \draw[-latex, thick] (l_\i) -- (la_\i) -| (oplus_\i);
        };
        
        \node[circle, line width=5pt, draw=gray!20, inner sep=2pt, minimum size=0pt] (max_filled) at (max) {\phantom{max}};
        \node[circle, thick, draw=black, text=black, fill=white, inner sep=2pt, minimum size=0pt] (max_label) at (max) {max};

        \foreach \i in {0, ..., \imgnumrows} {
            \draw[-, gray!20, line width=4pt] (fcl_\i) -- (midl_\i) -- (max);
            \draw[-latex, thick] (fcl_\i) -- (midl_\i) -- (max);
            
            \draw[-, gray!20, line width=4pt] (max) -- (midr_\i) -- (otimes_\i);
            \draw[-, thick] (max) -- (midr_\i) -- (otimes_\i);
        };
    \end{tikzpicture}};

    \foreach \i in {0, ..., \imgnumrows} {
        \node[bigo, left=2.5em of l_\i] (input_\i) {};
        \draw[-latex, thick] (input_\i) -- (background.west |- l_\i);
        \draw[-, thick] (background.west |- l_\i) -- (l_\i);

        \node[bigo, right=2.5em of fcr_\i] (output_\i) {};
        \draw[-latex, thick] (fcr_\i) -- (output_\i);
    };
\end{tikzpicture}
  \label{fig:model-architecture-2}
\end{subfigure}
\caption{
\textbf{PCT model architecture.} Diagram showing \textbf{Left}: The embedding network combining pre-existing conditions \protect\tikz\protect\draw[fill=BlueGreen!50] circle(0.8ex);, day offsets \protect\tikz\protect\draw[fill=NavyBlue!40] circle(0.8ex);, risk message clusters \protect\tikz\protect\draw[fill=Red!50] circle(0.8ex); \protect\tikz\protect\draw[fill=YellowGreen!50] circle(0.8ex);, and symptoms information \protect\tikz\protect\draw[fill=Dandelion!50] circle(0.8ex); to be fed into a stack of 5 either Deep-Set (DS) or Set Transformer (ST) blocks  \protect\tikz\protect\draw[fill=gray!20] circle(0.8ex);. \textbf{Right-top}: (ST) self-attention block featuring Multi-Head Dot Product Attention (MHDPA), Fully-Connected layers with ReLU (FC + R) and residual connections. \textbf{Right-bottom:} (DS) set processing block. Here, the $\otimes$ denotes concatenation and the $\oplus$ addition operation. %
}
\label{fig:model-architecture}
\end{figure}

To these ends, we construct a general architectural scaffold in which any neural network that maps between sets may be used. In this work, we experiment with Set Transformers \citep{lee2018set} and a variant of Deep Sets \citep{zaheer2017deep}. The former is a variant of Transformers \citep{vaswani2017attention}, which model pairwise interactions between all set elements via multi-head dot-product attention and can therefore be quite expressive, but scales quadratically with elements. The latter relies on iterative max-pooling of features along the elements of the set followed by a broadcasting of the aggregated features, and scales linearly with the number of set elements (see Figure~\ref{fig:model-architecture}).

In both cases, we first compute two categories of embeddings: per-day and per-encounter. Per-day, we have embedding MLP modules $\phi_{h}$ of health status, $\phi_{g}$ for the health profile, and a linear map $\phi_{\delta d}$ for the day-offset $\delta d = d - d'$. 
Per-encounter, the cluster matrix $e_i^d$ can be expressed as the set $\{(r^{d}_{j \to i}, n^{d}_{j \to i})\}_{j \in K_i^d}$ where $j$ is in the set $K_i^d$ of putative anonymous persons encountered by $i$ at $d$, $r_{j \to i}$ is the risk level sent from $j$ to $i$ and $n_{j \to i}$ is the number of repeated encounters between $i$ and $j$ at $d$. Accordingly, we define the risk level embedding function $\phi_e^{(r)}$ and $\phi_e^{(n)}$ for the number of repeated encounters. $\phi_e^{(r)}$ is parameterized by an embedding matrix, while $\phi_e^{(n)}$ by the vector 
\begin{align}
(\phi_{e}^{(n)}(n))_{2i} = \sin{\left(\nicefrac{n}{10000^i}\right)} \; \text{and} \; (\phi_{e}^{(n)}(n))_{2i + 1} = \cos{\left(\nicefrac{n}{10000^i}\right)}
\end{align}
which resembles the positional encoding in \citep{vaswani2017attention} and counteracts the spectral bias of downstream MLPs \citep{rahaman2019spectral,mildenhall2020nerf}. We tried alternative encoding schemes like thermometer encodings \citep{buckman2018thermometer}, but they did not perform better in our experiments. Next, we join the per-day and per-encounter embeddings to obtain $\mathcal{D}_i^d$ and $\mathcal{E}_i^d$ respectively. Where $\otimes$ is the concatenation operation, we have: 
\begin{align}
\label{eq:joined-embeddings}
\mathcal{D}_{i}^{d} &= \{ \phi_{h}(h_{i}^{d'}) \otimes \phi_{g}(g_i) \otimes \phi_{\delta d}(d' - d) \,|\, d' \in \{d,...,d-d_{max}\} \} \\
\mathcal{E}_{i}^{d} &= \{ \phi_{e}^{(r)}(r^{d'}_{j \rightarrow i}) \otimes \phi_{e}^{(n)}(n^{d'}_{j \rightarrow i}) \otimes \phi_{h}(h_{i}^{d'}) \otimes \phi_{\delta d}(d' - d) \,|\, j \in K_i^{d'}, \, \forall \, d' \in \{d,...,d-d_{max}\} \} \nonumber
\end{align}
The union of $\mathcal{D}_i^d$ and $\mathcal{E}_i^d$ forms the input to a set neural network $f_{S}$, that predicts infectiousness:
\begin{align}
\vec{\hat{y}}_i^d = f_{S}(\mathcal{O}_i^d) \text{ where } \mathcal{O}_i^d = \mathcal{D}_i^d \cup \mathcal{E}_i^d \text{ and } \vec{\hat{y}}_i^d \in \mathbb{R}_{+}^{d_{max}}
\end{align}

The trunk of both models -- deep-set (DS-PCT) and set-transformer (ST-PCT) -- is a sequence of 5 set processing blocks  (see Figure \ref{fig:model-architecture}). 
A subset of outputs from these blocks (corresponding to $\mathcal{D}_i^d$) is processed by a final MLP to yield $\mathbf{\hat y}_i^d$.
As a training objective for agent $i$, we minimize the Mean Squared Error (MSE) between $\vec{y}_i^{d}$ (which is generated by the simulator) and the prediction $\hat{\vec{y}}_i^d$. We treat each agent $i$ as a sample in batch to obtain the sample loss
$L_i = \text{MSE}(\vec{y}_i^d, \vec{\hat{y}}_i^d) = \frac{1}{d_{max}}\sum_{d'= d - d_{max}}^{d}(y_i^{d'}-\hat{y}_i^{d'})^2.$
The net loss is the sum of $L_i$ over all agents $i$. 
\subsection{Training Procedure}
\label{subsec:training-procedure}
Unlike existing contact tracing methods like BCT and the NHSx contact tracing app \citep{briers2020risk} that rely on simple and hand-designed heuristics to estimate the risk of infection, our core hypothesis is that methods using a machine-learning based predictor can learn from patterns in rich but noisy signals that might be available locally on a smartphone. 
In order to test this hypothesis prior to deployment in the real world, we require a simulator built with the objective to serve as a testbed for app-based contact tracing methods. We opt to use COVIsim \citep{gupta2020covisim}, which features an agent specific virology model together with realistic agent behaviour, mobility, contact and app-usage patterns. 
With the simulator, we generate large datasets of $\mathcal{O}(10^7)$ samples comprising the input and target variables defined in sections \ref{sec:problem_setup} and \ref{sec:distributed_inference}. Most importantly for our prediction task, this includes a continuous-valued infectiousness parameter as target for each agent. 

We use $240$ runs of the simulator to generate this dataset, where each simulation is configured with parameters randomly sampled from selected intervals (See Appendix \ref{app:training}). These parameters include the adoption rate of the contact tracing app, initial fraction of exposed agents in the population, the likelihood of agents ignoring the recommendations, and strength of social distancing and mobility restriction measures. 
There are two reasons for sampling over parameter ranges: First, the intervals these parameters are sampled from reflect both the intrinsic uncertainty in the parameters, as well as the practical setting and limitations of an app-based contact tracing method (e.g. we only care about realistic levels of app usage). Second, randomly sampling these key parameters significantly improves the diversity of the dataset, which in turn yields predictors that can be more robust when deployed in the real world. The overall technique resembles that of \textbf{domain-randomization} \citep{tobin2017domain, sadeghi2016cad2rl} in the Sim2Real literature, where its efficacy is well studied for transfer from RGB images to robotic control, e.g. \citep{chebotar2018closing, openai2018learning}.

We use $200$ runs for training and the remaining $40$ for validation (full training and other reproducibility details are in Appendix \ref{app:experimental_details}).
The model with the best validation score is selected for \emph{online} evaluation, wherein we embed it in the simulator to measure the reduction in the $R$ as a function of the average number of contacts per agent per day. While the evaluation protocol is described in section~\ref{sec:experiments}, we now discuss how we mitigate a fundamental challenge that we share with offline reinforcement learning methods, that of \textbf{auto-induced distribution shift} when the model is used in the simulation loop \citep{levine2020offline, hiads}.

To understand the problem, consider the case where the model is trained on simulation runs where PCT is driven by  ground-truth infectiousness values, i.e. an \emph{oracle} predictor. 
The predictions made by an oracle will in general differ from ones made by a trained model, as will resulting dynamics of spread of disease. This leads to a distribution over contacts and epidemiological scenarios that the model has not encountered during training. 
For example, oracle-driven PCT might even be successful in eliminating the disease in its early phases -- a scenario unlikely to occur in model-driven simulations. For similar reasons, oracle-driven simulations will also be less diverse than model-driven ones.  
To mitigate these issues, we adopt the following strategy: 
First, we generate an initial dataset with simulations driven by a \emph{noisy-oracle}, i.e. we add multiplicative and additive noise to the ground-truth infectiousness to partially emulate the output distribution resulting from trained models. The corresponding noise levels are subject to domain-randomization (as described above), resulting in a dataset with some diversity in epidemiological scenarios and contact patterns. Having trained the predictor on this dataset (until early-stopped), we generate another dataset from simulations driven by the thus-far trained predictor, in place of the noisy-oracle.
We then fine-tune the predictor on the new dataset (until early stopped) to obtain the final predictor. This process can be repeated multiple times, in what we call \textbf{iterative retraining}. We find  three steps yields a good trade-off between performance and compute requirement.

\label{sec:agent-based-epi-simulator}

\section{Experiments}
\label{sec:experiments}
We evaluate the proposed PCT methods, \textbf{ST-PCT }and \textbf{DS-PCT}, and benchmark them against: test-based BCT; a rule-based FCT \textbf{Heuristic} method proposed by \citet{gupta2020covisim}\footnote{While we did experiment with additional methods like linear regression and MLPs, they did not improve the performance over the rule-based FCT heuristic proposed by \citet{gupta2020covisim}.};
and a baseline \textbf{No Tracing (NT)} scenario which corresponds to recommendation level 1 (some social distancing).

\begin{figure}[htp]
\centering
\begin{subfigure}{.49\textwidth}
  \centering
  \includegraphics[width=1\linewidth]{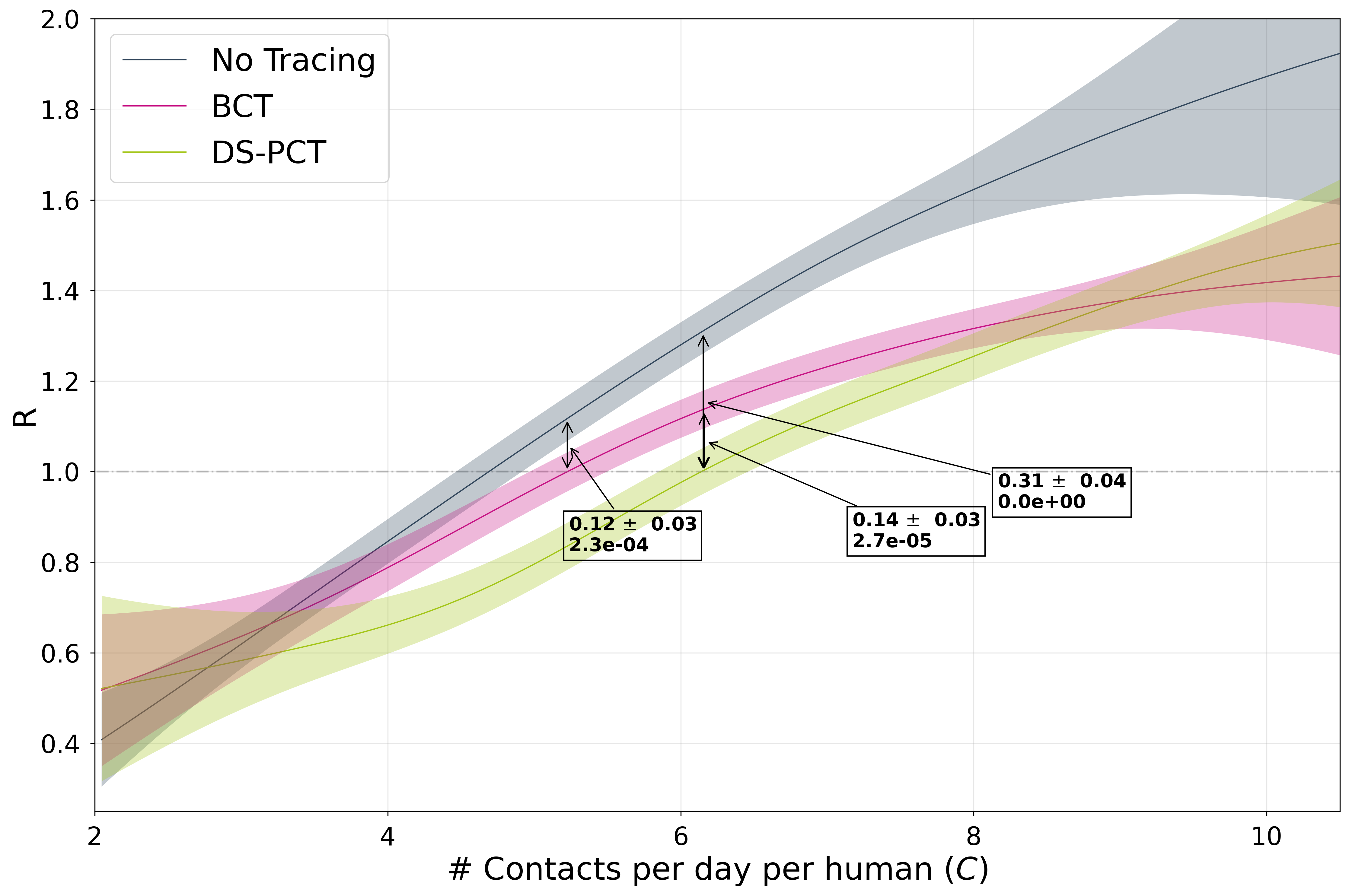}
  \label{fig:ds_pareto}
\end{subfigure}
\begin{subfigure}{.49\textwidth}
  \centering
  \includegraphics[width=1\linewidth]{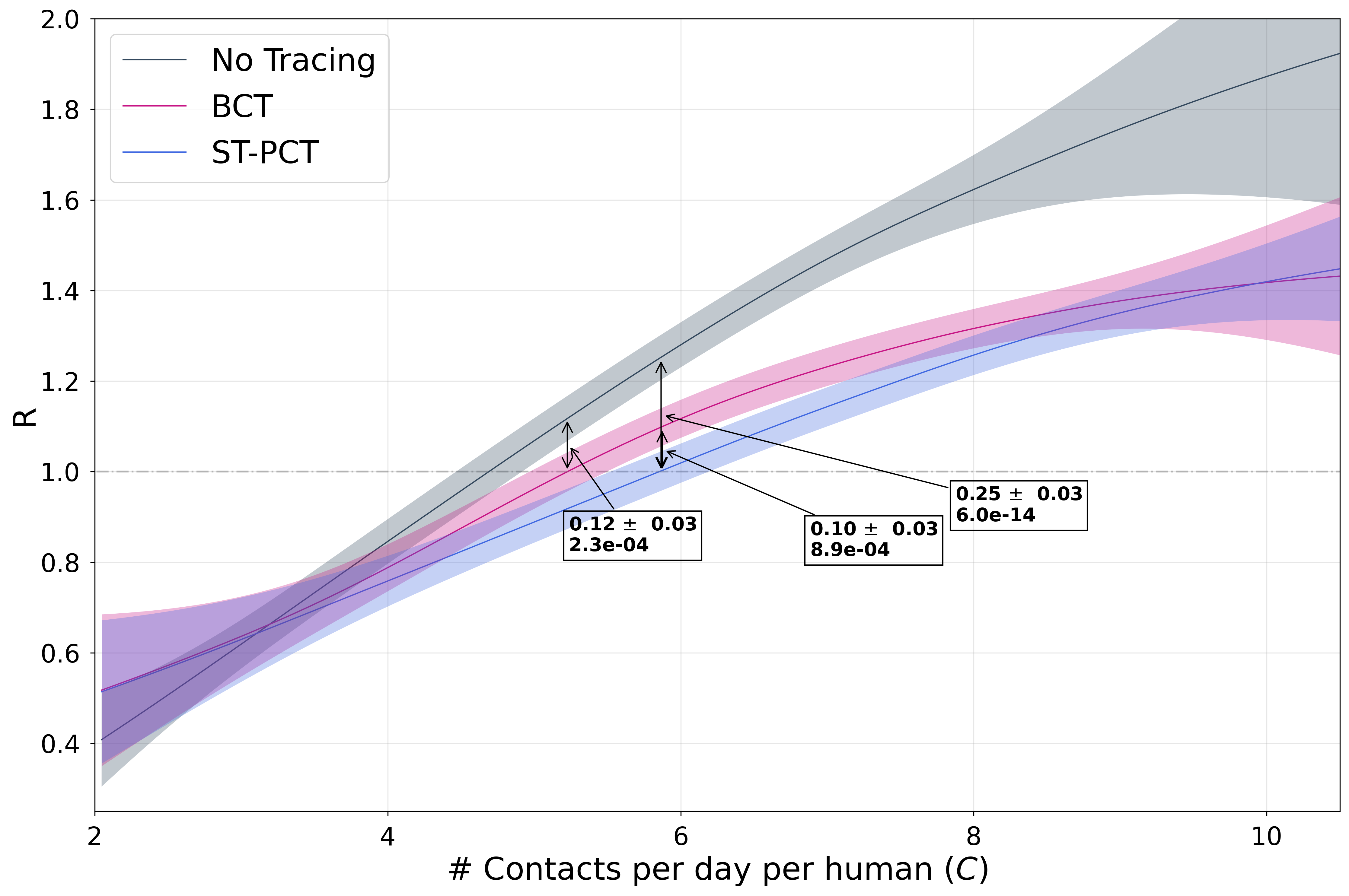}
  \label{fig:st_pareto}
\end{subfigure}
\caption{\textbf{Pareto frontier of mobility and disease spread:} reproduction number $R$ as a function of mobility. In boxes, average difference +/- standard error, p-value under null hypothesis of no difference. %
Note that small differences in $R$ over time produce large changes to the number of cases. 
\textbf{Gist:} All methods span a wide range of $R$ values in their Pareto frontier, including values for the No Tracing scenario which have $R$ below 1, achieved by imposing strong restrictions on mobility (e.g. a lockdown). 
However DCT methods are able to reduce $R$ at much lower cost to average mobility. Compared to NT, BCT has a 12\% advantage in $R$, DS-PCT (\textbf{left}) 31\% and ST-PCT 25\% (\textbf{right)}.
}
\label{fig:paretos}
\end{figure}

\textit{\underline{\textbf{EXP1}}:} In Figure~\ref{fig:paretos}, 
we plot the Pareto frontier between spread of disease ($R$) and the amount of restriction imposed. To traverse the frontier at 60\% adoption rate, we sweep through multiple values of the simulator's \textbf{global mobility scaling factor}, a parameter which controls the strength of social distancing and mobility restriction measures. 
Each method uses $3000$ agents for $50$ days with $12$ random seeds, and we plot the resulting number of contacts per day per human
We fit a Gaussian Process Regressor to the simulation outcomes and highlight the average reduction in $R$ obtained in the vicinity of $R \approx 1$, finding 
DS-PCT and ST-PCT reduce $R$ to below 1 at a much lower cost to mobility on average. 
For subsequent plots, we select the number of contacts per day per human such that the no-tracing baseline yields a realistic post-lockdown $R$ of around $1.2$ \citep{brisson2020}\footnote{To estimate this number, we use the GP regression fit in figure~\ref{fig:paretos} and consider the $x$ value for which the mean of the NT process is at $1.2$. We find an estimate at $5.61$, and select only the runs yielding effective number of contacts that lie within the interval $(5.61 - 0.5, 5.61 + 0.5)$.}.

\begin{figure}[htp]
\centering
\begin{subfigure}{.49\textwidth}
\includegraphics[width=1\linewidth]{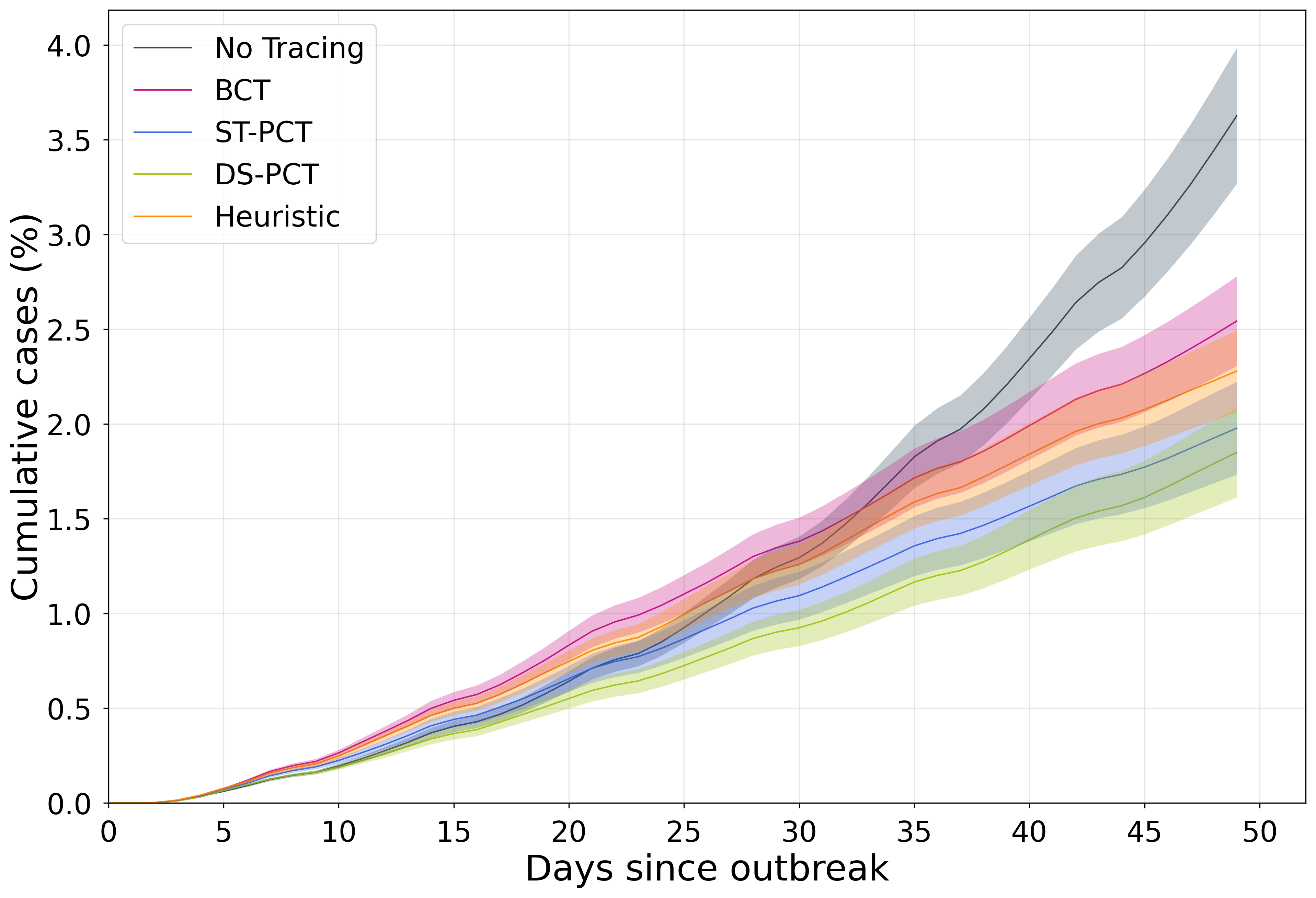}
\end{subfigure}
\begin{subfigure}{.49\textwidth}
\includegraphics[width=1\linewidth]{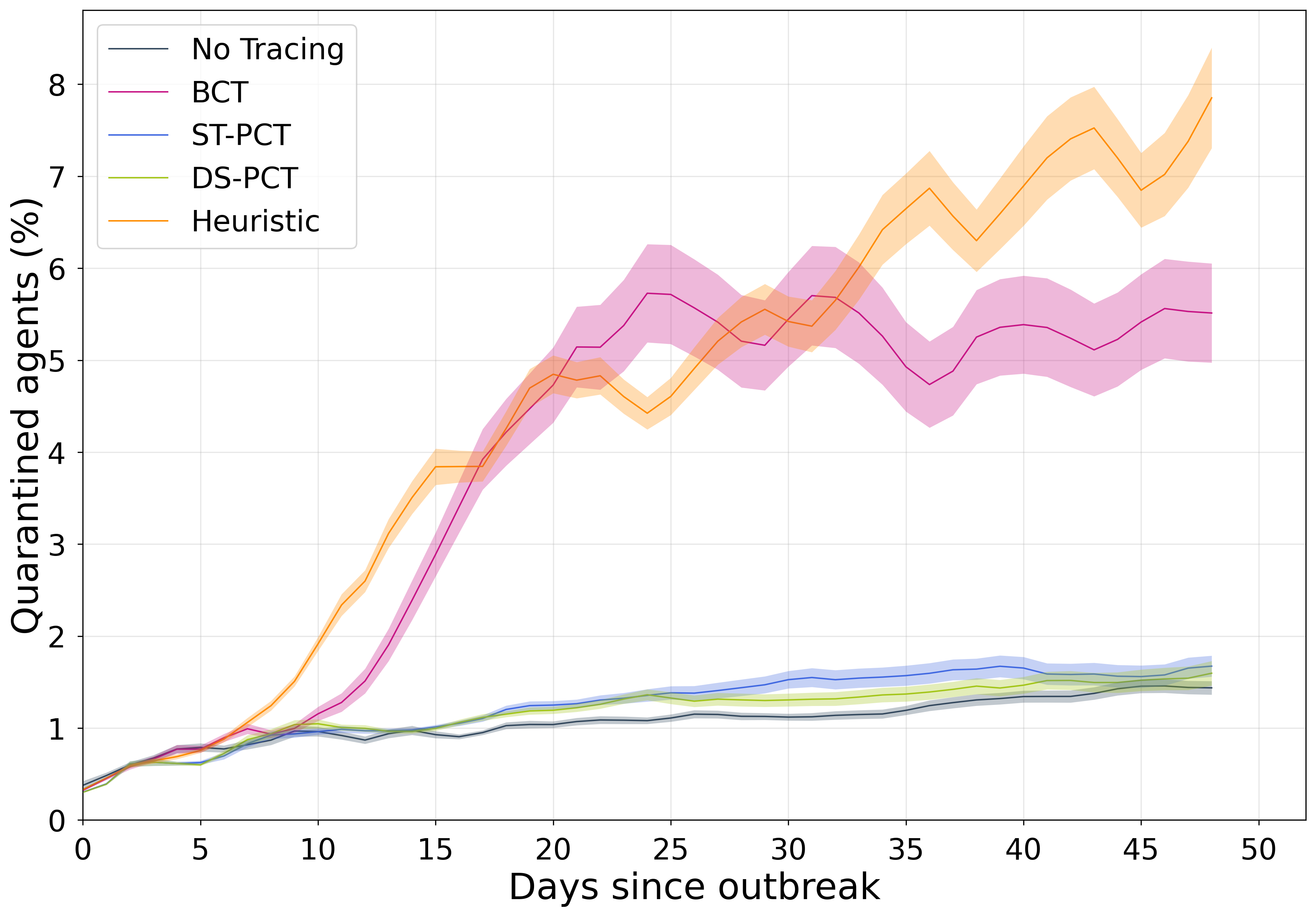}
\end{subfigure}
\caption{\textbf{Left:} Cumulative case counts for each method, 60\% adoption, 50 days, with all runs normalized to $5.61 \pm 0.5$ effective contacts per day per agent. \textbf{Right:} Mobility restriction for the same experiments (fraction of quarantines). \textbf{Gist:} For a similar number of cumulative cases to other DCT methods (left), PCT methods impose very little mobility restriction (right), close to NT.}
\label{fig:case-counts}    
\end{figure}

\textit{\underline{\textbf{EXP2}}:} Figure \ref{fig:case-counts} compares various DCT methods in terms of their cumulative cases and their fraction of \textbf{false quarantine}:  the number of healthy agents the method wrongly recommends to quarantine.
Again, all DCT methods have a clear advantage over the no-tracing baseline, while the number of agents wrongly recommended quarantine is much lower for ML-enabled PCT. \footnote{Note that the baseline no tracing method also has false quarantines, because household members of an infected individual are also recommended quarantine, irrespective of whether they are infected.} 

\begin{figure}[h!]
\centering
\begin{subfigure}{.49\textwidth}
\includegraphics[width=1\linewidth]{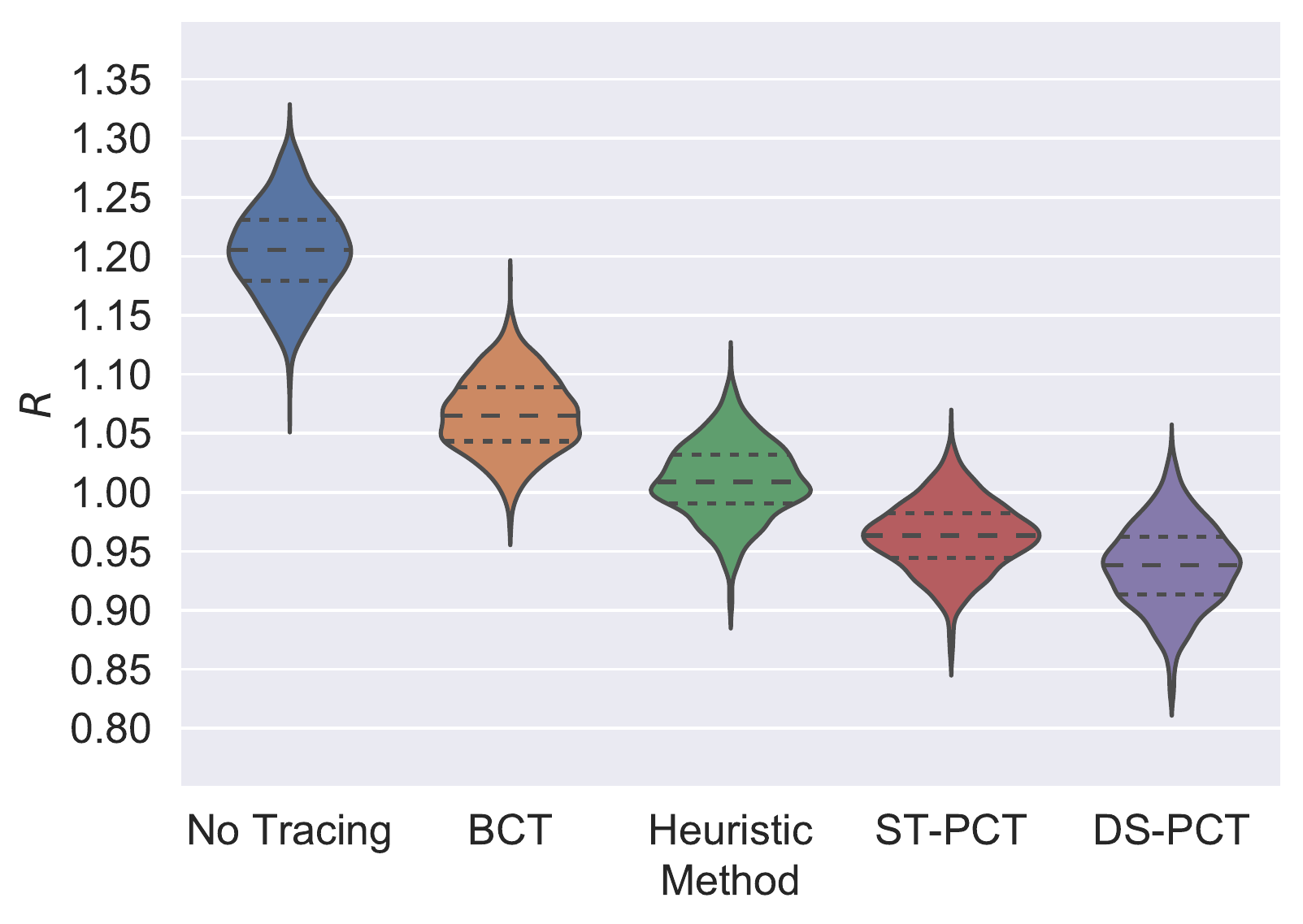}
\label{fig:_comparisons}    
\end{subfigure}
\begin{subfigure}{.49\textwidth}
\includegraphics[width=1\linewidth]{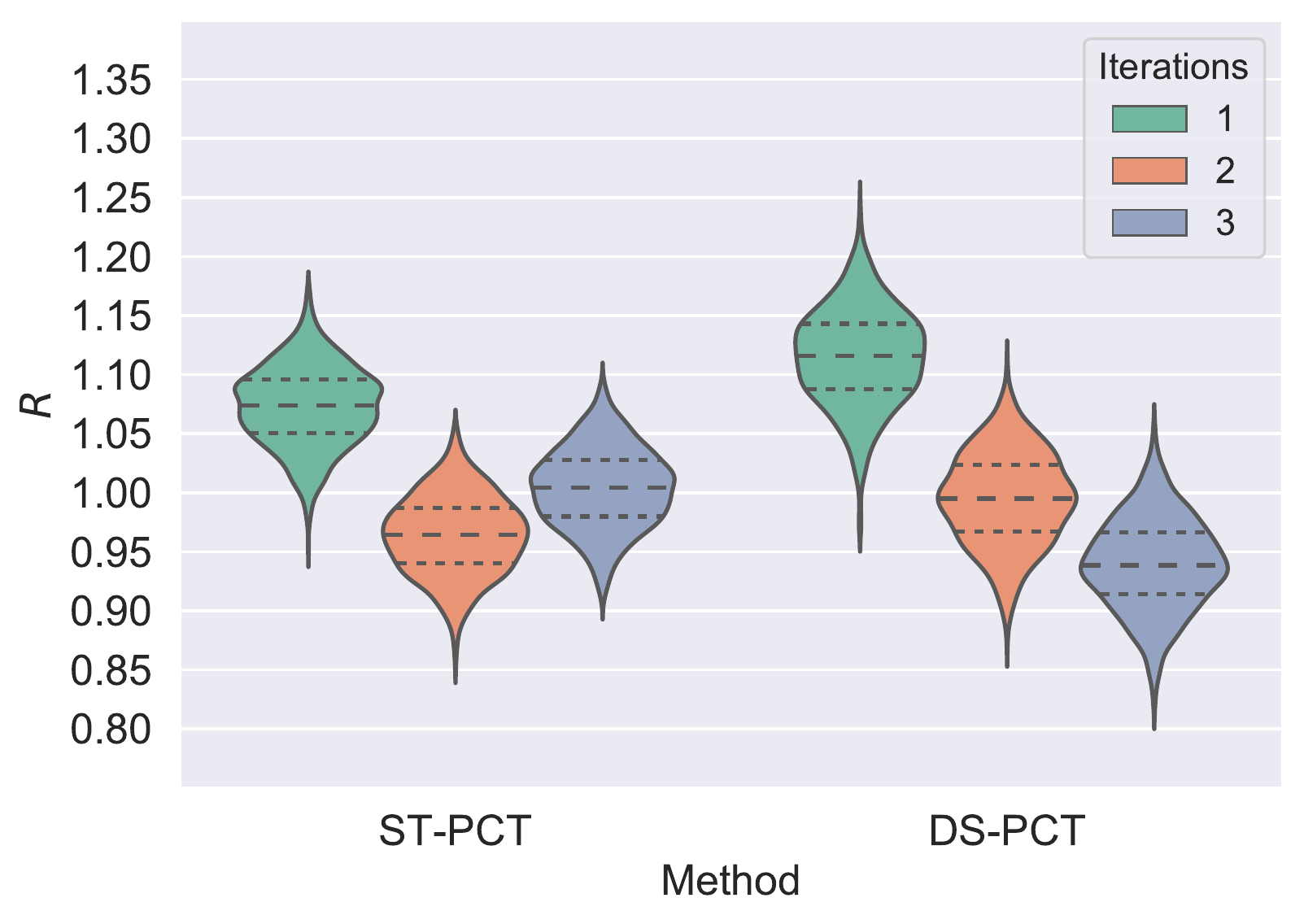}
\label{fig:_iterations}
\end{subfigure}
\caption{\textbf{Method Comparison and Retraining}. \textbf{Left:} Distribution (median and quartiles) of $R$ at 60\% adoption where the ML methods outperform the Heuristic and BCT methods proposed by \citet{gupta2020covisim}. \textbf{Right:} Iterative re-training significantly improves model performance. Upon the second re-training, however, it seems that ST-PCT begins to overfit. }
\label{fig:comparisons}
\end{figure}

\textit{\underline{\textbf{EXP3}}:} In Figure~\ref{fig:comparisons} \textbf{Left} we plot the bootstrapped distribution of mean $R$ for different DCT methods. Recall $R$ is an estimate of how many other agents an infectious agent will infects,
i.e. even a numerically small improvement in $R$ could yield an exponential improvement in the number of cases. 
We find that both PCT methods yield a clear improvement over BCT and the rule-based heuristic, all of which  significantly improve over the no-tracing baseline. We hypothesize this is because deep neural networks are better able to capture the non-linear relationship between features available on the phone, interaction patterns between agents, and individual infectiousness. 

\textit{\underline{\textbf{EXP4}}:} In Figure~\ref{fig:comparisons} \textbf{Right} we investigate the effect of \textbf{iterative retraining} on the machine-learning based methods, DS-PCT and ST-PCT. We evaluate all 3 iterations of each method in the same experimental setup as in Figure~\ref{fig:comparisons} and find that DS-PCT benefits from the 3 iterations, while ST-PCT saturates at the second iteration. We hypothesize that this is a form of overfitting, given that the set transformer (ST) models all-to-all interactions and is therefore more expressive than deep-sets (DS).

\begin{figure}[h ]
    {\caption{\textbf{Adoption rate comparison}. We compare  all methods for adoption rates between 0\% (NT) and 60\%. 
    \textbf{Gist:} All methods are able to improve over NT, even at low adoption rates. At 30\% and 45\%, ST-PCT performs the best by a relatively wide margin while DS-PCT outperforms it at 60\%. 
    }\label{fig:adoption}}
    {\includegraphics[width=0.95\textwidth]{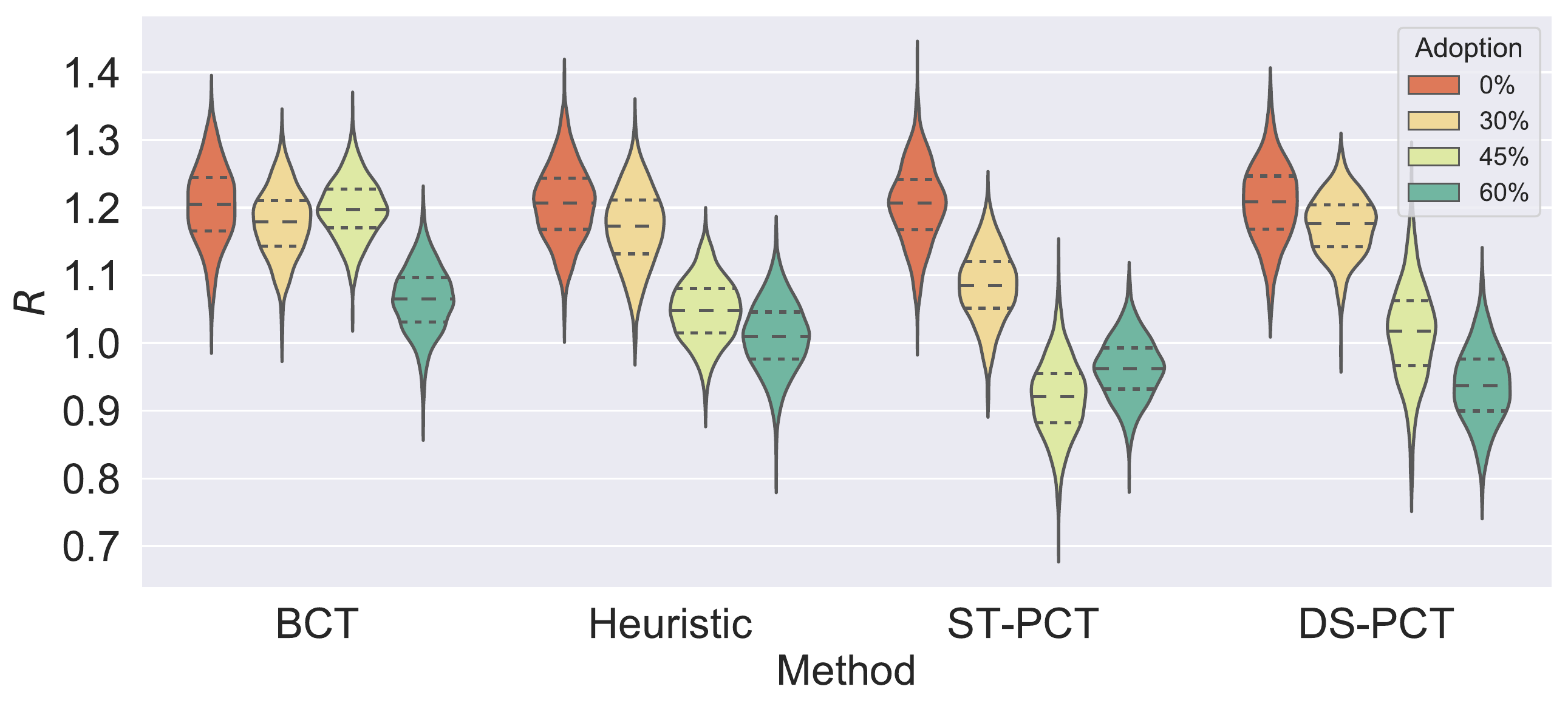}}
    
\end{figure}

\textbf{\textit{\underline{EXP5}}:}  In Figure~\ref{fig:adoption}, we analyze the sensitivity of the various methods to adoption rate, which measures what percent of the population actively uses the CT app. Adoption rate is an important parameter for DCT methods, as it directly determines the effectiveness of an app. We visualize the effect of varying the adoption rate on the reproduction number $R$. Similarly to prior work \citep{lowadoption}, we find that all DCT methods improve over the no-tracing baseline even at low adoptions
and PCT methods dominate at all levels of adoption.
    
 \FloatBarrier

\iffalse

\begin{figure}[htp!]
\centering
\includegraphics[width=0.28\columnwidth]{figuresPARETO_R_VALUES.png}
\caption{
Comparison of the trade-off  between 
Pareto frontier @Eilif. TODO discussion
}
\label{fig:risk_levels}
\end{figure}

\fi
\iffalse
\subsection{Sensitivity Analysis - App Adoption rate} \todo{need to show the relative effect of uptake
against binary tracing}
We perform sensitivity analysis of the best-performing trained Transformer as compared to Binary Contact Tracing, plotting performance 

See Figure \ref{fig:sensitivity}

\begin{figure}[htp!]
\centering
\includegraphics[width=.39\columnwidth]{sensitivity-adoption.png}
\caption{
Sensitivity analysis. TODO discussion
}
\label{fig:sensitivity}
\end{figure}
\fi

\section{Conclusion}
Our results demonstrate the potential benefit of digital contact tracing approaches for saving lives while reducing mobility restrictions and preserving privacy, independently confirming previous reports \citep{ferretti2020quantifying, lowadoption}.
Of all methods in our study,  %
we find that deep learning based PCT provides the best trade-off between restrictions on mobility and reducing the spread of disease under a range of settings, making it a potentially powerful tool for saving lives in a safe deconfinement.
This area of research holds many interesting avenues of future work in machine learning, including: (1) The comparison of methods for fitting parameters of the epidemiological simulator data using spatial information (not done here to avoid the need for GPS), (2) Using reinforcement learning to learn (rather than expert hand-tune) the mapping from estimated infectiousness and individual-level features (number of contacts, age, etc.)  %

Accurate and practical models for contact tracing are only a small part of the non-pharmaceutical efforts against the pandemic, a response to which is a complex venture necessitating cooperation between public health, government, individual citizens, and scientists of many kinds - epidemiologists, sociologists, behavioural psychologists, virologists, and machine learning researchers, among many others. We hope this work can play a role in fostering this necessary collaboration.

\section*{Broader Impact}
\label{sec:broader-impact}

This project is motivated by the goal of saving more lives than manual tracing or standard digital tracing which are both putting people in only two categories. This can be achieved by providing early graded warning of possible infectiousness, even before a person has symptoms, thanks to the combination of risk levels propagating across smartphones and the ability to predict those risk levels with machine learning. The potential to save lives is of huge benefit, but it must be appropriately weighed against the costs. This kind of mobile application raises serious questions about privacy, dignity and democracy which cannot be ignored. We contributed to a much longer paper which considers these social impact questions at length (a full citation will be provided in the non-anonymous version of the paper). See also~\citep{Bengio2020privacy} on the privacy of COVID-19
tracing apps, which is aligned with this work. We created a non-profit organization - a form of data trust - independent of governments and corporations to manage the project, with a board which includes representatives of civil society in addition to experts in public health, epidemiology, ethics, and machine learning, as well as representatives of governments. This organization would disappear (along with any data) at the end of the pandemic, and its sole mission is to protect the health, privacy and dignity of people in accordance with the principles in the Montreal Declaration for the Responsible Development of AI. 

Of note here is how these considerations have already imposed serious constraints on the information available to the risk predictor and training procedure (see Section \ref{sec:privacy_approach} above). Clearly, people who do not have a smartphone (around 15\% in our country) would not be able to receive recommendations. This might mean that they would not be able to help prevent others in their community from being infected by them, but they would benefit from a reduction in spread of the disease due to others who are equipped with a smartphone and using the app. There are also hardware makers building cheap Bluetooth devices (e.g. 5\$) which could be used by such people (think especially about older people and children, in particular, who might not be able to use a smartphone). Failing to convince enough users to adopt the app could mean that the privacy risks taken by those who use the app and share their data with the ML server might result in little collective benefit. To mitigate this, the proposed app provides a coarse hotspot map functionality to Public Health by aggregating risk statistics over a neighbourhood or small town. This would be useful even if a very small fraction of the population (like 10\%) were to use the app. Such a map would provide warnings of increasing risk in a neighborhood even before the outbreak translates into increased hospitalizations (because symptoms and contact information are often available long before a test result is obtained). Even at that low level of participation, the data collected would be of great value in terms of epidemiological modeling, i.e., to better understand the disease and its transmission, and thus guide public health policies.
Another social question of interest is whether an app like this would favor particular groups in society, for example, the most wealthy (because they have more smartphones and possibly more app uptake will be seen in their community). We propose that such issues be monitored by using census data to correlate prediction error with socio-economic status and race (on a per-neighborhood basis). If such disparities are found, then a mitigating strategy would be to reweigh (or oversample) examples from these disadvantaged communities in order to reduce the error bias.

\section*{Acknowledgements}

We thank all members of the broader COVI project \url{https://covicanada.org} for their dedication and teamwork; it was a pleasure and honour to work with such a great team. We also thank the Mila and Empirical Inference communities, in particular Xavier Bouthillier, Vincent Mai, Gabriele Prato, and Georgios Arvanitidis, for detailed and helpful feedback on early drafts.
The authors gratefully acknowledge the following funding sources: NSERC, IVADO, CIFAR.
This project could not have been completed without the resources of MPI-IS cluster, Compute Canada \& Calcul Quebec, in particular the Beluga cluster. In gratitude we have donated to a project to help understand and protect the St. Lawrence beluga whales for whom the cluster is named \url{https://baleinesendirect.org/}

\newpage
\bibliographystyle{iclr2021_conference}
\bibliography{refs}

\begin{thebibliography}{3}
\providecommand{\natexlab}[1]{#1}
\providecommand{\url}[1]{\texttt{#1}}
\expandafter\ifx\csname urlstyle\endcsname\relax
  \providecommand{\doi}[1]{doi: #1}\else
  \providecommand{\doi}{doi: \begingroup \urlstyle{rm}\Url}\fi

\bibitem[Bengio \& LeCun(2007)Bengio and LeCun]{Bengio+chapter2007}
Yoshua Bengio and Yann LeCun.
\newblock Scaling learning algorithms towards {AI}.
\newblock In \emph{Large Scale Kernel Machines}. MIT Press, 2007.

\bibitem[Goodfellow et~al.(2016)Goodfellow, Bengio, Courville, and
  Bengio]{goodfellow2016deep}
Ian Goodfellow, Yoshua Bengio, Aaron Courville, and Yoshua Bengio.
\newblock \emph{Deep learning}, volume~1.
\newblock MIT Press, 2016.

\bibitem[Hinton et~al.(2006)Hinton, Osindero, and Teh]{Hinton06}
Geoffrey~E. Hinton, Simon Osindero, and Yee~Whye Teh.
\newblock A fast learning algorithm for deep belief nets.
\newblock \emph{Neural Computation}, 18:\penalty0 1527--1554, 2006.

\end{thebibliography}


\begin{thebibliography}{48}
\providecommand{\natexlab}[1]{#1}
\providecommand{\url}[1]{\texttt{#1}}
\expandafter\ifx\csname urlstyle\endcsname\relax
  \providecommand{\doi}[1]{doi: #1}\else
  \providecommand{\doi}{doi: \begingroup \urlstyle{rm}\Url}\fi

\bibitem[Abueg et~al.(2020)Abueg, Hinch, Wu, Liu, Probert, Wu, Eastham, Shafi,
  Rosencrantz, Dikovsky, Cheng, Nurtay, Abeler-D{\"o}rner, Bonsall, McConnell,
  O{\textquoteright}Banion, and Fraser]{lowadoption}
Matthew Abueg, Robert Hinch, Neo Wu, Luyang Liu, William J~M Probert, Austin
  Wu, Paul Eastham, Yusef Shafi, Matt Rosencrantz, Michael Dikovsky, Zhao
  Cheng, Anel Nurtay, Lucie Abeler-D{\"o}rner, David~G Bonsall, Michael~V
  McConnell, Shawn O{\textquoteright}Banion, and Christophe Fraser.
\newblock Modeling the combined effect of digital exposure notification and
  non-pharmaceutical interventions on the covid-19 epidemic in washington
  state.
\newblock \emph{medRxiv}, 2020.
\newblock \doi{10.1101/2020.08.29.20184135}.
\newblock URL
  \url{https://www.medrxiv.org/content/early/2020/09/02/2020.08.29.20184135}.

\bibitem[Aleta et~al.(2020)Aleta, Martin-Corral, y~Piontti, Ajelli, Litvinova,
  Chinazzi, Dean, Halloran, Longini~Jr, Merler, et~al.]{aleta2020modeling}
Alberto Aleta, David Martin-Corral, Ana~Pastore y~Piontti, Marco Ajelli, Maria
  Litvinova, Matteo Chinazzi, Natalie~E Dean, M~Elizabeth Halloran, Ira~M
  Longini~Jr, Stefano Merler, et~al.
\newblock Modeling the impact of social distancing, testing, contact tracing
  and household quarantine on second-wave scenarios of the covid-19 epidemic.
\newblock \emph{medRxiv}, 2020.

\bibitem[Alsdurf et~al.(2020)Alsdurf, Belliveau, Bengio, Deleu, Gupta,
  Ippolito, Janda, Jarvie, Kolody, Krastev, Maharaj, Obryk, Pilat, Pisano,
  Prud'homme, Qu, Rahaman, Rish, Rousseau, Sharma, Struck, Tang, Weiss, and
  Yu]{whitepaper}
Hannah Alsdurf, Edmond Belliveau, Yoshua Bengio, Tristan Deleu, Prateek Gupta,
  Daphne Ippolito, Richard Janda, Max Jarvie, Tyler Kolody, Sekoul Krastev,
  Tegan Maharaj, Robert Obryk, Dan Pilat, Valerie Pisano, Benjamin Prud'homme,
  Meng Qu, Nasim Rahaman, Irina Rish, Jean-Francois Rousseau, Abhinav Sharma,
  Brooke Struck, Jian Tang, Martin Weiss, and Yun~William Yu.
\newblock Covi white paper.
\newblock \emph{arXiv preprint arXiv:2005.08502}, 2020.

\bibitem[Baker et~al.(2020)Baker, Biazzo, Braunstein, Catania, Dall'Asta,
  Ingrosso, Krzakala, Mazza, Mézard, Muntoni, Refinetti, Mannelli, and
  Zdeborová]{baker2020probabilistic}
Antoine Baker, Indaco Biazzo, Alfredo Braunstein, Giovanni Catania, Luca
  Dall'Asta, Alessandro Ingrosso, Florent Krzakala, Fabio Mazza, Marc Mézard,
  Anna~Paola Muntoni, Maria Refinetti, Stefano~Sarao Mannelli, and Lenka
  Zdeborová.
\newblock Epidemic mitigation by statistical inference from contact tracing
  data, 2020.

\bibitem[Bengio et~al.(2020)Bengio, Janda, Yu, Ippolito, Jarvie, Pilat, Struck,
  Krastev, and Sharma]{Bengio2020privacy}
Yoshua Bengio, Richard Janda, Yun~William Yu, Daphne Ippolito, Max Jarvie, Dan
  Pilat, Brooke Struck, Sekoul Krastev, and Abhinav Sharma.
\newblock The need for privacy with public digital contact tracing during the
  covid-19 pandemic.
\newblock \emph{The Lancet Digital Health}, 2020.

\bibitem[Bergstra \& Bengio(2012)Bergstra and Bengio]{JMLR:v13:bergstra12a}
James Bergstra and Yoshua Bengio.
\newblock Random search for hyper-parameter optimization.
\newblock \emph{Journal of Machine Learning Research}, 13\penalty0
  (10):\penalty0 281--305, 2012.
\newblock URL \url{http://jmlr.org/papers/v13/bergstra12a.html}.

\bibitem[Briers et~al.(2020)Briers, Charalambides, and Holmes]{briers2020risk}
Mark Briers, Marcos Charalambides, and Chris Holmes.
\newblock Risk scoring calculation for the current nhsx contact tracing app.
\newblock \emph{arXiv preprint arXiv:2005.11057}, 2020.

\bibitem[Brisson \& et~al.(2020)Brisson and et~al.]{brisson2020}
Marc Brisson and et~al.
\newblock Épidémiologie et modélisation de l’évolution de la covid-19 au
  québec.
\newblock \emph{INSPQ}, 2020.
\newblock URL \url{https://www.inspq.qc.ca/covid-19/}.

\bibitem[Buckman et~al.(2018)Buckman, Roy, Raffel, and
  Goodfellow]{buckman2018thermometer}
Jacob Buckman, Aurko Roy, Colin Raffel, and Ian Goodfellow.
\newblock Thermometer encoding: One hot way to resist adversarial examples.
\newblock In \emph{International Conference on Learning Representations}, 2018.
\newblock URL \url{https://openreview.net/forum?id=S18Su--CW}.

\bibitem[Chebotar et~al.(2018)Chebotar, Handa, Makoviychuk, Macklin, Issac,
  Ratliff, and Fox]{chebotar2018closing}
Yevgen Chebotar, Ankur Handa, Viktor Makoviychuk, Miles Macklin, Jan Issac,
  Nathan Ratliff, and Dieter Fox.
\newblock Closing the sim-to-real loop: Adapting simulation randomization with
  real world experience, 2018.

\bibitem[El~Emam et~al.(2011)El~Emam, Jonker, Arbuckle, and
  Malin]{el2011systematic}
Khaled El~Emam, Elizabeth Jonker, Luk Arbuckle, and Bradley Malin.
\newblock A systematic review of re-identification attacks on health data.
\newblock \emph{PloS one}, 6\penalty0 (12):\penalty0 e28071, 2011.

\bibitem[Fan et~al.(2016)Fan, Li, and Heller]{fan}
Kai Fan, Chunyuan Li, and Katherine Heller.
\newblock A unifying variational inference framework for hierarchical
  graph-coupled hmm with an application to influenza infection.
\newblock In \emph{Proceedings of the Thirtieth AAAI Conference on Artificial
  Intelligence}, AAAI'16, pp.\  3828–3834. AAAI Press, 2016.

\bibitem[Ferretti et~al.(2020)Ferretti, Wymant, Kendall, Zhao, Nurtay,
  Abeler-D{\"o}rner, Parker, Bonsall, and Fraser]{ferretti2020quantifying}
Luca Ferretti, Chris Wymant, Michelle Kendall, Lele Zhao, Anel Nurtay, Lucie
  Abeler-D{\"o}rner, Michael Parker, David Bonsall, and Christophe Fraser.
\newblock Quantifying sars-cov-2 transmission suggests epidemic control with
  digital contact tracing.
\newblock \emph{Science}, 368\penalty0 (6491), 2020.
\newblock URL \url{https://doi.org/10.1126/science.abb6936}.

\bibitem[Gandhi et~al.(2020)Gandhi, Yokoe, and Havlir]{gandhi2020asymptomatic}
Monica Gandhi, Deborah~S. Yokoe, and Diane~V. Havlir.
\newblock Asymptomatic transmission, the achilles’ heel of current strategies
  to control {COVID}-19.
\newblock \emph{New England Journal of Medicine}, 382\penalty0 (22):\penalty0
  2158--2160, 2020.
\newblock URL \url{https://doi.org/10.1056/NEJMe2009758}.

\bibitem[Grantz et~al.(2020)Grantz, Lee, McGowan, Lee, Metcalf, Gurley, and
  Lessler]{grantz2020maximizing}
Kyra~H Grantz, Elizabeth~C Lee, Lucy~D'Agostino McGowan, Kyu~Han Lee,
  C~Jessica~E Metcalf, Emily~S Gurley, and Justin Lessler.
\newblock Maximizing and evaluating the impact of test-trace-isolate programs.
\newblock \emph{medRxiv}, 2020.

\bibitem[Gupta et~al.(2020)Gupta, Maharaj, Weiss, Rahaman, Alsdurf, Sharma,
  Minoyan, Harnois-Leblanc, Schmidt, Charles, Deleu, Williams, Patel, Qu,
  Bilaniuk, Caron, Carrier, Ortiz-Gagne, Rousseau, Buckeridge, Ghosn, Zhang,
  Sch\"olkopf, Tang, Rish, Pal, Merckx, Muller, and Bengio]{gupta2020covisim}
Prateek Gupta, Tegan Maharaj, Martin Weiss, Nasim Rahaman, Hannah Alsdurf,
  Abhinav Sharma, Nanor Minoyan, Soren Harnois-Leblanc, Victor Schmidt,
  Pierre-Luc~St. Charles, Tristan Deleu, Andrew Williams, Akshay Patel, Meng
  Qu, Olexa Bilaniuk, G\'aetan~Marceau Caron, Pierre-Luc Carrier, Satya
  Ortiz-Gagne, Marc-Andre Rousseau, David Buckeridge, Joumana Ghosn, Yang
  Zhang, Bernhard Sch\"olkopf, Jian Tang, Irina Rish, Christopher Pal, Joanna
  Merckx, Eilif~B. Muller, and Yoshua Bengio.
\newblock {COVI}sim: an agent-based model for evaluating methods of digital
  contact tracing, 2020.

\bibitem[{Health Canada}(2020)]{canada-public-health-prolonged-exposure}
{Health Canada}.
\newblock Public health management.
\newblock
  https://www.canada.ca/en/public-health/services/diseases/2019-novel-coronavirus-infection/health-professionals/interim-guidance-cases-contacts.html,
  April 2020.

\bibitem[Hellewell et~al.(2020)Hellewell, Abbott, Gimma, Bosse, Jarvis,
  Russell, Munday, Kucharski, Edmunds, Sun, et~al.]{hellewell2020feasibility}
Joel Hellewell, Sam Abbott, Amy Gimma, Nikos~I Bosse, Christopher~I Jarvis,
  Timothy~W Russell, James~D Munday, Adam~J Kucharski, W~John Edmunds, Fiona
  Sun, et~al.
\newblock Feasibility of controlling covid-19 outbreaks by isolation of cases
  and contacts.
\newblock \emph{The Lancet Global Health}, 2020.

\bibitem[Heneghan et~al.(2020)Heneghan, Brassey, and
  Jefferson]{heneghan2020sars}
Carl Heneghan, Jon Brassey, and Tom Jefferson.
\newblock Sars-cov-2 viral load and the severity of {COVID}-19, 2020.
\newblock URL
  \url{https://www.cebm.net/covid-19/sars-cov-2-viral-load-and-the-severity-of-covid-19/}.
\newblock Accessed: 2020-06-02.

\bibitem[Herbrich et~al.(2020)Herbrich, Rastogi, and Vollgraf]{CRISP}
Ralf Herbrich, Rajeev Rastogi, and Roland Vollgraf.
\newblock Crisp: A probabilistic model for individual-level covid-19 infection
  risk estimation based on contact data, 2020.

\bibitem[Hinch et~al.(2020)Hinch, Probert, Nurtay, Kendall, Wymant, Hall, and
  Fraser]{hinch2020effective}
Robert Hinch, W~Probert, A~Nurtay, M~Kendall, C~Wymant, Matthew Hall, and
  C~Fraser.
\newblock Effective configurations of a digital contact tracing app: A report
  to nhsx.
\newblock \emph{en. In:(Apr. 2020). Available here. url: https://github.
  com/BDI-pathogens/covid-19\_instant\_tracing/blob/master/Report}, 2020.

\bibitem[JJ et~al.(2013)JJ, ST, R, J, NT, PC, WD, A, DD, A, H, T, and
  DS.]{FRED}
Grefenstette JJ, Brown ST, Rosenfeld R, Depasse J, Stone NT, Cooley PC, Wheaton
  WD, Fyshe A, Galloway DD, Sriram A, Guclu H, Abraham T, and Burke DS.
\newblock Fred (a framework for reconstructing epidemic dynamics): An
  open-source software system for modeling infectious diseases and control
  strategies using census-based populations.
\newblock \emph{BMC Public Health}, 2013.

\bibitem[Krueger et~al.(2020)Krueger, Maharaj, and Leike]{hiads}
David Krueger, Tegan Maharaj, and Jan Leike.
\newblock Hidden incentives for auto-induced distributional shift.
\newblock \emph{arXiv preprint arXiv:2009.09153}, 2020.

\bibitem[Lee et~al.(2018)Lee, Lee, Kim, Kosiorek, Choi, and Teh]{lee2018set}
Juho Lee, Yoonho Lee, Jungtaek Kim, Adam~R. Kosiorek, Seungjin Choi, and
  Yee~Whye Teh.
\newblock Set transformer: A framework for attention-based
  permutation-invariant neural networks, 2018.
\newblock URL \url{https://arxiv.org/abs/1810.00825}.

\bibitem[Levine et~al.(2020)Levine, Kumar, Tucker, and Fu]{levine2020offline}
Sergey Levine, Aviral Kumar, George Tucker, and Justin Fu.
\newblock Offline reinforcement learning: Tutorial, review, and perspectives on
  open problems.
\newblock \emph{arXiv preprint arXiv:2005.01643}, 2020.

\bibitem[Li et~al.(2020)Li, Wang, Dong, Wang, Huang, Xu, and Xia]{li2020false}
Dasheng Li, Dawei Wang, Jianping Dong, Nana Wang, He~Huang, Haiwang Xu, and
  Chen Xia.
\newblock False-negative results of real-time reverse-transcriptase polymerase
  chain reaction for severe acute respiratory syndrome coronavirus 2: role of
  deep-learning-based ct diagnosis and insights from two cases.
\newblock \emph{Korean journal of radiology}, 21\penalty0 (4):\penalty0
  505--508, 2020.

\bibitem[Lokhov et~al.(2014)Lokhov, M{\'e}zard, Ohta, and
  Zdeborov{\'a}]{lokhov2014inferring}
Andrey~Y Lokhov, Marc M{\'e}zard, Hiroki Ohta, and Lenka Zdeborov{\'a}.
\newblock Inferring the origin of an epidemic with a dynamic message-passing
  algorithm.
\newblock \emph{Physical Review E}, 90\penalty0 (1), 2014.
\newblock URL \url{https://doi.org/10.1103/PhysRevE.90.012801}.

\bibitem[Lorch et~al.(2020)Lorch, Kremer, Trouleau, Tsirtsis, Szanto,
  Schölkopf, and Gomez-Rodriguez]{lorch2020spatiotemporal}
Lars Lorch, Heiner Kremer, William Trouleau, Stratis Tsirtsis, Aron Szanto,
  Bernhard Schölkopf, and Manuel Gomez-Rodriguez.
\newblock Quantifying the effects of contact tracing, testing, and containment,
  2020.

\bibitem[Mildenhall et~al.(2020)Mildenhall, Srinivasan, Tancik, Barron,
  Ramamoorthi, and Ng]{mildenhall2020nerf}
Ben Mildenhall, Pratul~P Srinivasan, Matthew Tancik, Jonathan~T Barron, Ravi
  Ramamoorthi, and Ren Ng.
\newblock Nerf: Representing scenes as neural radiance fields for view
  synthesis.
\newblock \emph{arXiv preprint arXiv:2003.08934}, 2020.

\bibitem[Minka(2001)]{Minka2001_Expectation_Propagation}
Thomas~P. Minka.
\newblock Expectation propagation for approximate bayesian inference.
\newblock In \emph{Proceedings of the Seventeenth Conference on Uncertainty in
  Artificial Intelligence}, UAI'01, pp.\  362–369, San Francisco, CA, USA,
  2001. Morgan Kaufmann Publishers Inc.
\newblock ISBN 1558608001.

\bibitem[Murphy(2020)]{KevinModel}
Kevin Murphy.
\newblock Bayesian contact tracing.
\newblock "Personal Correspondance.", 2020.

\bibitem[Murphy et~al.(1999)Murphy, Weiss, and Jordan]{murphy1999loopy}
Kevin~P Murphy, Yair Weiss, and Michael~I Jordan.
\newblock Loopy belief propagation for approximate inference: an empirical
  study.
\newblock In \emph{Proceedings of the Fifteenth conference on Uncertainty in
  artificial intelligence}, pp.\  467--475, 1999.

\bibitem[OpenAI et~al.(2018)OpenAI, Andrychowicz, Baker, Chociej, Jozefowicz,
  McGrew, Pachocki, Petron, Plappert, Powell, Ray, Schneider, Sidor, Tobin,
  Welinder, Weng, and Zaremba]{openai2018learning}
OpenAI, Marcin Andrychowicz, Bowen Baker, Maciek Chociej, Rafal Jozefowicz, Bob
  McGrew, Jakub Pachocki, Arthur Petron, Matthias Plappert, Glenn Powell, Alex
  Ray, Jonas Schneider, Szymon Sidor, Josh Tobin, Peter Welinder, Lilian Weng,
  and Wojciech Zaremba.
\newblock Learning dexterous in-hand manipulation, 2018.

\bibitem[Rahaman et~al.(2019)Rahaman, Baratin, Arpit, Draxler, Lin, Hamprecht,
  Bengio, and Courville]{rahaman2019spectral}
Nasim Rahaman, Aristide Baratin, Devansh Arpit, Felix Draxler, Min Lin, Fred
  Hamprecht, Yoshua Bengio, and Aaron Courville.
\newblock On the spectral bias of neural networks.
\newblock In \emph{International Conference on Machine Learning}, pp.\
  5301--5310. PMLR, 2019.

\bibitem[Russell(2020)]{russell2020rise}
A~Russell.
\newblock The rise of coronavirus hate crimes.
\newblock \emph{The New Yorker}, 2020.

\bibitem[Sadeghi \& Levine(2016)Sadeghi and Levine]{sadeghi2016cad2rl}
Fereshteh Sadeghi and Sergey Levine.
\newblock Cad2rl: Real single-image flight without a single real image.
\newblock \emph{arXiv preprint arXiv:1611.04201}, 2016.

\bibitem[Satorras \& Welling(2020)Satorras and Welling]{satorras2020neural}
Victor~Garcia Satorras and Max Welling.
\newblock Neural enhanced belief propagation on factor graphs, 2020.

\bibitem[Stevens(2020)]{stevens2020}
Harry Stevens.
\newblock Why outbreaks like coronavirus spread exponentially, and how to
  "flatten the curve", 2020.
\newblock URL
  \url{https://www.washingtonpost.com/graphics/2020/world/corona-simulator/}.

\bibitem[Sweeney(2002)]{10.1142/S0218488502001648}
Latanya Sweeney.
\newblock K-anonymity: A model for protecting privacy.
\newblock \emph{Int. J. Uncertain. Fuzziness Knowl.-Based Syst.}, 10\penalty0
  (5):\penalty0 557–570, October 2002.
\newblock ISSN 0218-4885.
\newblock \doi{10.1142/S0218488502001648}.
\newblock URL \url{https://doi.org/10.1142/S0218488502001648}.

\bibitem[Tobin et~al.(2017)Tobin, Fong, Ray, Schneider, Zaremba, and
  Abbeel]{tobin2017domain}
Josh Tobin, Rachel Fong, Alex Ray, Jonas Schneider, Wojciech Zaremba, and
  Pieter Abbeel.
\newblock Domain randomization for transferring deep neural networks from
  simulation to the real world, 2017.

\bibitem[Vaswani et~al.(2017)Vaswani, Shazeer, Parmar, Uszkoreit, Jones, Gomez,
  Kaiser, and Polosukhin]{vaswani2017attention}
Ashish Vaswani, Noam Shazeer, Niki Parmar, Jakob Uszkoreit, Llion Jones,
  Aidan~N. Gomez, Lukasz Kaiser, and Illia Polosukhin.
\newblock Attention is all you need, 2017.

\bibitem[Verity et~al.(2020)Verity, Okell, Dorigatti, Winskill, Whittaker,
  Imai, Cuomo-Dannenburg, Thompson, Walker, Fu, et~al.]{verity}
Robert Verity, Lucy~C Okell, Ilaria Dorigatti, Peter Winskill, Charles
  Whittaker, Natsuko Imai, Gina Cuomo-Dannenburg, Hayley Thompson, Patrick~GT
  Walker, Han Fu, et~al.
\newblock Estimates of the severity of coronavirus disease 2019: a model-based
  analysis.
\newblock \emph{The Lancet infectious diseases}, 2020.

\bibitem[Vespignani et~al.(2020)Vespignani, Chinazzi, Davis, Mu, y~Piontti,
  Samay, Gioannini, Litvinova, Milano, Paolotti, Quaggiotto, Rossi, Tizzoni,
  and Vismara]{gleam}
Alessandro Vespignani, Matteo Chinazzi, Jessica~T. Davis, Kunpeng Mu,
  Ana~Pastore y~Piontti, Nicole Samay, Xinyue Xiongand~Corrado Gioannini, Maria
  Litvinova, Paolo Milano, Daniela Paolotti, Marco Quaggiotto, Luca Rossi,
  Michele Tizzoni, and Ivan Vismara.
\newblock Gleam project.
\newblock \url{https://www.gleamproject.org/}, 2020.
\newblock Accessed: 2020-06-11.

\bibitem[Winn \& Bishop(2005)Winn and Bishop]{winn2005variational}
John Winn and Christopher~M Bishop.
\newblock Variational message passing.
\newblock \emph{Journal of Machine Learning Research}, 6\penalty0
  (Apr):\penalty0 661--694, 2005.

\bibitem[Wood et~al.(2020)Wood, Warrington, Naderiparizi, Weilbach, Masrani,
  Harvey, Scibior, Beronov, and Nasseri]{wood2020planning}
Frank Wood, Andrew Warrington, Saeid Naderiparizi, Christian Weilbach, Vaden
  Masrani, William Harvey, Adam Scibior, Boyan Beronov, and Ali Nasseri.
\newblock Planning as inference in epidemiological models.
\newblock \emph{arXiv preprint arXiv:2003.13221}, 2020.

\bibitem[Woodhams(2020)]{DCT-list}
Samuel Woodhams.
\newblock Covid-19 digital rights tracker - contact tracing apps.
\newblock
  \url{https://docs.google.com/spreadsheets/d/1_BCKlMuniEhzvpQ-ha0jhdksvqdINUAUHA8J9LSr_Dc/edit#gid=0},
  2020.
\newblock Accessed: 2020-06-11.

\bibitem[Yedidia et~al.(2000)Yedidia, Freeman, and
  Weiss]{yedidia2001generalized}
Jonathan~S. Yedidia, William~T. Freeman, and Yair Weiss.
\newblock Generalized belief propagation.
\newblock In \emph{Proceedings of the 13th International Conference on Neural
  Information Processing Systems}, pp.\  668–674, 2000.
\newblock URL \url{https://dl.acm.org/doi/10.5555/3008751.3008848}.

\bibitem[Zaheer et~al.(2017)Zaheer, Kottur, Ravanbakhsh, Poczos, Salakhutdinov,
  and Smola]{zaheer2017deep}
Manzil Zaheer, Satwik Kottur, Siamak Ravanbakhsh, Barnabas Poczos, Russ~R
  Salakhutdinov, and Alexander~J Smola.
\newblock Deep sets.
\newblock In \emph{Advances in neural information processing systems}, pp.\
  3391--3401, 2017.

\end{thebibliography}

\newpage
\section*{Appendix 1: Glossary of bolded terms} \label{app:glossary}

\paragraph{\textbf{Contact tracing}} Finding the people who have been in contact with an infected person, typically using information such as test results and phone surveys, and recommending that they quarantine themselves to prevent further spread of the disease.

\paragraph{\textbf{Manual contact tracing} } Method for contact tracing using trained professionals  who interview diagnosed individuals to identify people that have come into contact with an infected person and recommend a change in their behavior (or further testing).

\paragraph{\textbf{Digital contact tracing (DCT)}} Predominantly smartphone-based methods of identifying individuals at high risk of contracting an infectious disease based on their interactions, mobility patterns, and available medical information such as test results.

\paragraph{\textbf{Binary contact tracing (BCT)}} Methods which use binary information (e.g. tested positive or negative) to perform contact tracing, thus putting users in binary risk categories (at-risk or not-at-risk). 

\paragraph{\textbf{Proactive contact tracing (PCT)}} DCT methods which produce graded risk information to mitigate spread of disease, thus putting users in graded risk categories (more-at-risk or less-at-risk), potentially using non-binary clues like reported symptoms which are available earlier than test results.

\paragraph{\textbf{Agent-based epidemiological models (ABMs)} } Also called individual-based, this type of simulation defines rules of behaviour for individual agents, and by stepping forward in time, contact patterns between agents are generated in a ``bottom up'' fashion. Contrast with \textbf{Population-level epidemiological models}.

\paragraph{\textbf{Risk of infection} } Expected infectiousness, or probability of infection for a susceptible agent by an infectious agent given a qualifying contact (15+ minutes at under 2 meters).

\paragraph{\textbf{Contact}} An encounter between 2 agents which lasts at least 15 minutes with a distance under 2 meters \citep{canada-public-health-prolonged-exposure}.

\paragraph{\textbf{Risk Level}} A version of the risk of infection quantized into 16 bins (for
reducing the number of bits being exchanged,
for better privacy protection). The thresholds are selected by running the domain randomization with separate seeds and grouping risk messages such that there are an approximately uniform number in each bin.

\paragraph{\textbf{Risk mapping}} A table which maps floating point risk scalars into one of 16 discrete risk levels. 

\paragraph{\textbf{Recommendation mapping}} A table of which recommendation level should be associated with a given risk level.
Each recommendation level comes with a series
of specific recommendations
(e.g. "Limit contact with others", "wash hands frequently", "wear a mask when near others", 
"avoid public transportation", etc) which should be shown for a given risk level. Reminders of these recommendations may be sent via push notification more or less often depending
on the recommendation level.

\paragraph{\textbf{Population-level epidemiological models}} This type of model fits global statistics of a population with a mathematical model, typically sets of ordinary differential equations. The equations track some statistics over time, typically counts of people in each of a few ``compartments'', e.g. the number of Infected and Recovered, and the parameters of the equations are often tuned to match statistics collected from real data. It does not give any information about the individual-level contact patterns which
give rise to these statistics.

\paragraph{\textbf{Compartment model (SEIR)} } A compartment model tracks the counts of agents in each of several mutually-exclusive categories called ``compartments''. In an SEIR model, the 4 compartments are \textbf{Susceptible, Exposed, Infectious, and Recovered} (see entries for each of these words).

\paragraph{\textbf{Susceptible}} At risk of catching disease, but not infected.

\paragraph{\textbf{Exposed or infected}} Infected with the disease, (i.e. potentially carries some \textbf{viral load}).

\paragraph{\textbf{Infectious}} Carries sufficient viral load to transmit the disease to others. In real life, this typically means that the virus has multiplied sufficiently to overwhelm the immune system.

\paragraph{\textbf{Recovered}} No longer carries measurable viral load, after having been
infected. In real life this is typically measured by two successive negative \textbf{lab tests}.

\paragraph{\textbf{Viral load}} Viral load is the number of actual viral RNA in a person as measured by a \textbf{lab test}.

\paragraph{\textbf{Effective viral load}} A term we introduce representing a number between 0 and 1 which we use as a proxy for viral load. It could be converted to an actual viral load via multiplying by the maximum amount of viral RNA detectable by a lab test.

\paragraph{\textbf{Infectiousness}} Degree to which an agent is able to transmit viruses to another agent. A factor in determining whether an encounter between a susceptible and an infectious agent results in the susceptible agent being exposed (it is a property of the infector agent; this does not consider environmental factors or the other agent's susceptibility; see \textbf{Transmission probability}).

\paragraph{\textbf{Attack rate}}  The proportion of individuals in a population who are infected over some time period.

\paragraph{\textbf{Contamination duration} } How long a location remains infectious after an infected person has visited it and shed viruses there.

\paragraph{\textbf{Transmission probability} } The chance that one agent infects another during an encounter. \textbf{Infectiousness} of the infected agent, \textbf{susceptibility} of the other agent, both their behaviour (e.g. usage of masks) and other environmental factors all play a role in determining this probability. 

\paragraph{\textbf{PCR Test}} Also, colloquially referred to as a lab test. In this context, this means a Polymerase Chain Reaction (PCR) test used to amplify viral DNA samples, typically obtained via a nasopharangeal swab.

\paragraph{\textbf{Asymptomatic}} When an infected person shows no symptoms. For COVID-19, many people remain asymptomatic for the entire course of their disease.

\paragraph{\textbf{Incubation days} }  The number of days from exposure (infection) until an agent shows symptoms. In real life this is the average amount of time it takes the viral RNA to multiply sufficiently in / burst out of host cells and cause an immune response, which is what produces the symptoms.

\paragraph{\textbf{Reproduction number ($R$)} } The average number of other agents an agent infects, measured over a certain window of time. We follow \citet{gupta2020covisim} and approximate $R$ by computing the infection tree and taking the ratio of the number of infected children divided by the number of parents who are recovered infectors.

\paragraph{\textbf{Serial interval} } The average number of days between successive symptom onsets (i.e. between the \textbf{incubation} time of the infector and the infectee).

\paragraph{\textbf{Generation time}} The number of days between successive infections (i.e. between the exposure of the infector and the infectee). Generation time is difficult
to measure directly so one can estimate it
from the {\bf serial interval}.

\paragraph{\textbf{Domain randomization}} A method of generating a training dataset by sampling from several distributions, each
corresponding to a different setting of
parameters of a simulator.
 This allows to generate a dataset which covers a potentially wide range of settings, improving generalization to environments which
 may not be well covered by the simulator
 in any one particular setting chosen
 a priori. 

\paragraph{\textbf{Oracle Predictor} } Baseline for the \textbf{Set Transformer} which uses ground-truth infectiousness levels as "predictions". Provides an expected upper-bound performance for the predictors and can be used
to provide risk level inputs when pre-training a
predictors.

\paragraph{\textbf{Multi-Layer Perceptron (MLP)} } A risk prediction model which concatenates all inputs (after heuristically aggregating to a fixed size vector the set-valued inputs), feeds these through several fully-connected neural network layers and is trained to predict infectiousness. See Section \ref{app:architectures}.

\paragraph{\textbf{Set Transformer (ST) Model} } A risk prediction model which uses a modified set transformer to process and attend to inputs, and is trained to predict infectiousness. See Section \ref{app:architectures}.

\paragraph{\textbf{Deep Set (DS) Model}} Similar structure as the Set Transformer, but using pooling instead of self-attention. This allows it to use less memory and compute as compared to the Set Transformer.

\paragraph{\textbf{Heuristic (HCT) Model} } A rule-based method for predicting risk histories and corresponding current behavior levels developed in \citet{gupta2020covisim}.

\paragraph{\textbf{Adoption rate} } The proportion of the total population which uses a digital contact tracing application.

\paragraph{\textbf{Domain randomization} } The technique of varying the parameters of a simulation environment to create a broad distribution of training data.

\paragraph{\textbf{Re-identification} } The process of matching anonymous data with publicly available data so as to determine which person is the owner.

\paragraph{\textbf{Big brother attacks} } "Big brother" is a reference to the 1984 book by George Orwell where a highly regulated government surveilled and controlled its citizenry. The term "big brother attacks" references the idea that an organized group (e.g., state, federal, academic or financial institution) may try to gain control of your data.

\paragraph{\textbf{Little brother attacks} } Little brother attacks references the idea that potential security threats are also posed by smaller actors like individuals, criminals, and data brokers. 

\paragraph{\textbf{Vigilante attacks} } Extra-judicial violence by an individual. In the context of this paper, specifically as a result of a recommendation shown to the user or gained through other means.

\paragraph{\textbf{Degree of Restriction} } Ratio between the number of people given a recommendation in the highest level of restriction (i.e. quarantine) relative to other categories.

\paragraph{\textbf{Global mobility scaling factor} } A simulator parameter enabling us to vary the amount of contacts which may lead to infection in the simulator. It allows us to scale the mobility to simulate pre-lockdown or post-lockdown environments.

\paragraph{\textbf{Auto-induced distribution shift} } A change in distribution of data observed by an agent or algorithm as a result of the agent or algorithm's actions.

\paragraph{\textbf{Iterative re-training} } The process of generating data using some method in a simulator, training a model on this data in a supervised manner, then evaluating the model in the simulator to produce more data.

\paragraph{\textbf{Pareto frontier} } The Pareto frontier is a way of evaluating a tradeoffs in a set of policies and environments with multi-dimensional outputs (e.g., viral spread and mobility). 

\newpage

\section*{Appendix 2: Motivating Example}
\begin{figure}[ht]
    \centering
    \begin{adjustbox}{width=\textwidth,center}
    \definecolor{ProgressDarkBlue}{RGB}{46, 57, 83}
\definecolor{ProgressNavyBlue}{RGB}{50, 79, 157}
\definecolor{ProgressBlue}{RGB}{92, 149, 204}
\definecolor{ProgressCyan}{RGB}{59, 172, 187}
\definecolor{ProgressGreen}{RGB}{59, 178, 134}
\definecolor{ProgressLightGreen}{RGB}{186, 189, 77}
\definecolor{ProgressYellow}{RGB}{252, 196, 62}

\begin{tikzpicture}[day/.style={rectangle, text=black, text width=60pt, text depth=80pt, text height=1em, minimum height=80pt, minimum width=60pt, inner sep=10pt, outer sep=0, font=\fontfamily{phv}\selectfont}, header/.style={day, inner sep=10pt, text width=80pt, text=white, font=\bfseries\Large\fontfamily{phv}\selectfont}, dayofweek/.style={rectangle, text width=60pt, text height=1em, minimum width=60pt, inner sep=10pt, outer sep=0, font=\bfseries\Large\fontfamily{phv}\selectfont, text=ProgressNavyBlue}]
    \newcommand{\scenario}{,,Jim has a contact with high-risk stranger at the grocery store,,Stranger starts\\showing symptoms,,Stranger's symptoms grow worse,Jim\\\textcolor{white}{\textbf{GOES}}\\to work,{Stranger sees doctor, gets tested},Test result comes back positive,,,Jim is\\contacted directly by public health}

    \node[header, fill=ProgressNavyBlue] (day0) at (0, 0) {Manual\\tracing\\only};
    \foreach \x [count=\i, remember=\i as \lasti (initially 0)] in \scenario {
    \ifnum \numexpr \i > 12
      \colorlet{DayFillColor}{ProgressYellow}
    \else
      \colorlet{DayFillColor}{ProgressBlue}
    \fi
    \node[day, fill=DayFillColor, right=2pt of day\lasti] (day\i) {\x};
    }
    \node[day, fill=none, right=2pt of day13, text width=10pt, minimum width=10pt] (day14) {};
    \fill[ProgressYellow] (day14.{south west}) -- (day14.{south east}) arc (-90:90:55pt) -- (day14.{north west}) -- cycle;
    
    \node[dayofweek, above=5pt of day1] (dayname0) {M};
    \newcommand{\daysofweek}{T,W,T,F,S,S,M,T,W,T,F,S,S}
    \foreach \y [count=\i, remember=\i as \lasti (initially 0)] in \daysofweek {
    \node[dayofweek, right=2pt of dayname\lasti] (dayname\i) {\y};
    }
    \draw[line width=2pt, ProgressNavyBlue] (dayname0.{north west}) -- (dayname6.30);
    \draw[line width=2pt, ProgressNavyBlue] (dayname7.{north west}) -- (dayname13.30);

    \newcommand{\scenariobinary}{Jim installs the app,,Jim has a contact with high-risk stranger at the grocery store,,Stranger starts\\showing symptoms,,Stranger's symptoms grow worse,Jim\\\textcolor{white}{\textbf{GOES}}\\to work,{Stranger sees doctor, gets tested},Test result comes back positive,,,Jim is\\contacted directly by public health}

    \node[header, fill=ProgressNavyBlue, below=10pt of day0] (binary0) {Binary\\contact\\tracing};
    \foreach \x [count=\i, remember=\i as \lasti (initially 0)] in \scenariobinary {
    \ifnum \numexpr \i > 8
      \colorlet{DayFillColor}{ProgressYellow}
    \else
      \colorlet{DayFillColor}{ProgressBlue}
    \fi
    \node[day, fill=DayFillColor, right=2pt of binary\lasti] (binary\i) {\x};
    }
    \node[day, fill=none, right=2pt of binary13, text width=10pt, minimum width=10pt] (binary14) {};
    \fill[ProgressYellow] (binary14.{south west}) -- (binary14.{south east}) arc (-90:90:55pt) -- (binary14.{north west}) -- cycle;

    \newcommand{\scenarioours}{Jim installs the app,,Jim has a contact with high-risk stranger at the grocery store,,Stranger starts\\showing symptoms,,Stranger's symptoms grow worse,Jim\\\textcolor{white}{\textbf{DOES NOT}}\\go to work,{Stranger sees doctor, gets tested},Test result comes back positive,,,Jim is\\contacted directly by public health}

    \node[header, fill=ProgressDarkBlue, below=10pt of binary0] (ours0) {Our\\approach};
    \foreach \x [count=\i, remember=\i as \lasti (initially 0)] in \scenarioours {
    \ifnum \numexpr \i < 9
      \ifnum \numexpr \i > 6
        \colorlet{DayFillColor}{ProgressLightGreen}
      \else
      \ifnum \numexpr \i > 4
        \colorlet{DayFillColor}{ProgressGreen}
      \else
        \ifnum \numexpr \i > 2
          \colorlet{DayFillColor}{ProgressCyan}
        \else
          \colorlet{DayFillColor}{ProgressBlue}
        \fi
      \fi
      \fi
    \else
      \colorlet{DayFillColor}{ProgressYellow}
    \fi
    \node[day, fill=DayFillColor, right=2pt of ours\lasti] (ours\i) {\x};
    }
    \node[day, fill=none, right=2pt of ours13, text width=10pt, minimum width=10pt] (ours14) {};
    \fill[ProgressYellow] (ours14.{south west}) -- (ours14.{south east}) arc (-90:90:55pt) -- (ours14.{north west}) -- cycle;
\end{tikzpicture}
    \end{adjustbox}
    \caption{\textbf{Motivating example comparing manual, binary, and proactive contact tracing}: This example shows the potential effectiveness of early warnings in controlling the spread of the infection. Manual tracing is delayed because of the time
    between diagnosis and calling all contacts. Both manual and digital contact
    tracing are sending late signals because they only make use of the strongest
    possible signal (positive diagnosis). The proposed ML approach takes advantage
    of reported symptoms and the propagation of risk signals between phones to
    obtain much earlier signals.}
    \label{fig:early-awareness-example}
\end{figure}

\section*{Appendix 3: Comparison of approach to related work}\label{sec:rel-work}
\noindent{\bf Epidemiological Modeling and Simulations.} Given that modeling contact-tracing requires capturing past interactions, it is mathematically complicated to consider the dynamic nature of network interactions.
An agent-based model (simulator), on the other hand, gives us maximum flexibility to incorporate real data and/or assumptions easily
and emulate the effect of personalized
policies (such as resulting from the proposed
app).
Models with binary contagion and random-walk mobility are ubiquitous  \citep{stevens2020}.
The appeal of such models is their simplicity; they are easy to code, fast to run, and can give a general picture of some aspects of disease spread. There are other models with increasing complexity of either the mobility model, contagion model, agent demographics, or some combination of these, e.g. \citep{verity,gleam,wood2020planning, lorch2020spatiotemporal,hinch2020effective}.

GLEAM (Global Epidemic And Mobility model) \citep{gleam} is an off-the-shelf simulation platform for epidemics which offers mobility patterns and demographic information, and uses a generic format for defining how a disease depends on these two things.
Similarly, FRED (Framework for Reconstructing Epidemiological Dynamics) provides an open-source, agent-based model with realistic social networks and US demographics. \citet{FRED}, and \citep{wood2020planning} build an intervention-planning tool on top of this simulator. These works are comparable to our simulator only; they do not do any contact tracing or individual risk prediction. 

The work presented in \citep{lorch2020spatiotemporal} and \citep{ferretti2020quantifying} is closely related to ours.
A detailed mobility model for interactions is presented  in \citep{lorch2020spatiotemporal}, but the epidemiological model of the disease is much simpler there.
First, the contact graph is built based on their mobility model. The next step is an  implementation of various policy interventions by health authorities, which includes contact tracing. This is done instead of  building a contact graph that takes into account policy interventions and contact tracing apps at an individual level. Next, most similarly to our work, \citep{ferretti2020quantifying} proposes an app-based tracing and recommendations. 
However, their epidemiological model is a very simple differential-equation   model, and interventions for controlling the disease are computed analytically.

\noindent{\bf Risk estimation approaches.} While the applications deployed so far are based primarily on binary contact tracing as discussed above,   some probabilistic risk estimation approaches, similar to ours, have been developed for other diseases or applications, and have begun to be applied to COVID-19.  For example,   \citep{baker2020probabilistic} uses the susceptible-infected-recovered (SIR) model and describes the dynamical process of infection propagation using the dynamical message passing equations from \citep{lokhov2014inferring};   the probability of each node (person) to be in a specific state (S,I or R) is estimated via the dynamic message-passing (DMP) algorithm, which
belongs to the family of  local message-passing methods similar to belief propagation (BP) algorithm  \citep{yedidia2001generalized}  for  estimating marginal probability distributions over the network nodes; despite BP being only an approximate inference method,   not  guaranteed to converge to the correct marginals when the underlying graphical model has (undirected) cycles, it  demonstrated remarkable performance in various applications. A similar approach based on BP was also discussed in \citep{KevinModel}. 
Furthermore, there are recent extensions of belief propagation approach graph neural nets \citep{satorras2020neural}. Also, another recent work uses Gibbs sampling and SEIR model for their test-trace-isolate approach \citep{CRISP}.
However, such approaches typically rely on the   knowledge of the social  interaction graph, which is not available in our case due to privacy and security constraints.

To the best of our knowledge, our work is the first to use an approach based on  detailed agent-based epidemiological model together with a model of phone app messaging to generate simulated data for training an ML-based predictor of individual-level risk.

\iffalse %
\section*{Appendix: Epidemiology }
\subsection{Virology}

Details of effective viral load curve; parameters and where they come from.
Explanation of immune response and 

\subsection{Symptoms}

Details about how symptoms relate to viral load curve, and where we get stats for frequencies of each symptom. Include refs to progression of flu and ref that says symptoms look more like flu than other SARS

\subsection{Implementation details of Simulator}
@Tegan

\section*{Appendix: Demographics and other Statistics}
\subsection{Social mixing}
@ Gaetan @Prateek

\subsection{Age, Biological sex, and Comorbidities}

\fi

\textbf{Digital contact tracing for COVID-19}
\citep{hinch2020effective, hellewell2020feasibility, aleta2020modeling, grantz2020maximizing} study, either via simulations or mathematical models, the conditions under which BCT can be effective. Toward addressing the issues with BCT, ~\citep{hinch2020effective} show in simulation that using self-reported symptoms in addition to test results can greatly help control an outbreak.
Probabilistic (non-binary) approaches to the problem of contact tracing (e.g. \citet{baker2020probabilistic,satorras2020neural, briers2020risk}) typically assume full access to location histories and contact graph, an unacceptable violation of privacy in most places in the world. As a result, these methods are most often used for predicting overall patterns of disease spread.

\textbf{Agent-Based Models as a generative process}
Our use of an agent-based simulator \citep{gupta2020covisim} as a generative model allows us to generate fine-grained (continuous) values for expected infectiousness with realistic contact patterns. 
Most works performing probabilistic inference for disease modeling use a simple differential equation generative model, which does not characterize individual behaviour but rather the dynamics of transition between each of several mutually exclusive disease states.
Such models make many simplifying assumptions, such as contact patterns based on random walks, which make them unsuitable for individual-level prediction of infectiousness; they are typically used instead to infer latent variables such as infectiousness that would be consistent with the population-level statistics generated by the differential equation model, and/or to predict statistics of spread of the disease in a population, e.g. (ref, ref).
While agent-based models are widely used in epidemiological literature to model the spread of disease (see e.g. (ref) for review), to our knowledge we are the first to use an ABM as a generative model for training a deep learning-based infectiousness predictor. 

\textbf{Distributed inference and belief propagation}
Belief propagation in graphical models is often used for disease spread modeling, e.g. \citet{fan}. Some recent works have applied this to COVID-19; for example, \citet{baker2020probabilistic} use the susceptible-infected-recovered (SIR) model and describe the process of infection propagation using the dynamical message passing equations from \citet{lokhov2014inferring}. %
A work concurrent with ours follows a similar justification for modeling a latent parameter of expected infectiousness, using an SEIR model with inference via \citep{CRISP}. However, these approaches rely on a centralized social graph or a large number of bits exchanged between nodes, which is challenging both in terms of privacy and bandwidth. This challenge motivated our particular form of distributed inference where we pretrain the predictor and do not assume that the messages exchanged are probability distributions, but instead just informative input to the node-level predictor.

\section*{Appendix 4: Experimental details} \label{app:experimental_details}

\subsection*{ML architectures and baseline details:} \label{app:architectures}

\paragraph{Binary contact tracing} %
quarantines app-users who had high-risk encounters with an app-user who receives a positive PCR test. Under BCT1, if Alice gets a positive test result, then every user who encountered Alice within 14 days of her receiving the positive test result is sent a message which places them in app-recommendation level 3 (quarantine) for 14 days. Formally, $\zeta_i^d = \psi(\hat y_i^d) = 3$.

\paragraph{Set Transformer} 
Recall that in Section~\ref{sec:methodology-for-infectiousness-estimation} we proposed two parameterizations of the model (DS and ST in Figure \ref{fig:model-architecture}). In the first proposal, we use a set transformer (ST) to model interactions between the elements in the set $\mathbb{D}_i^d \cup \mathbb{E}_i^d$. We now describe the precise architecture of the model used. 

The model comprises 5 embedding modules, namely: the health status embedding $\phi_{hs}$, health profile embedding $\phi_{hp}$, day offset embedding $\phi_{do}$, risk message embedding $\phi_{e}^{(r)}$ and an embedding $\phi_{e}^{(n)}$ of the number of repeated encounters. 

The model was trained for 160 epochs on a domain randomized dataset (see below) comprising $\sim 10^7$ samples. We used a batch-size of 1024, resulting in $\sim 80k$ training steps. The learning rate schedule is such that the first $2.5k$ steps are used for linear learning-rate warmup, wherein the learning rate is linearly increased from $0$ to $2 \times 10^{-4}$, followed by a cosine annealing schedule that decays the learning rate from $2 \times 10^{-4}$ to $8 \times 10^{-6}$ in $50k$ steps.

\subsection*{Domain randomization} \label{app:training}

Inspired by research in hyper-parameter search \citep{JMLR:v13:bergstra12a} and recent advances in deep reinforcement learning \citep{tobin2017domain} we created the transformer's training data by sampling uniformly in the following ranges:

\iffalse
\begin{enumerate}
    \item App adoption rate $\in [0.3; 0.7]$
    \item People's general carefulness (which affects their probability of infection, their diligence in reporting potential symptoms, propensity to wear masks etc.) $\in [0.5; 0.8]$
    \item People's daily probability of not following recommendations \citep{ferretti2020quantifying} $\in [0; 0.04]$
    \item Proportion of the population initially exposed to COVID-19 $\in [0.0015; 0.011]$
    \item 
\end{enumerate}
\fi
\begin{enumerate}
    \item Adoption rate $\in [30 - 60]$
    \item Carefulness $\in [0.5 - 0.8]$ 
    \item Initial proportion of exposed people $\in [0.002, 0.006]$
    \item Oracle additive noise $\in [0.05 - 0.15]$
    \item Oracle multiplicative noise $\in [0.2 - 0.8]$
    \item Global mobility scaling factor $\in [0.3 - 0.9]$
    \item Symptoms dropout: likelihood of not reporting some symptoms $\in [0.1, 0.6]$
    \item Symptoms drop-in: Likelihood of falsely reporting symptoms $\in [0.0001, 0.001]$
    \item Quarantine dropou (test)t $\in [0.01, 0.03]$: likelihood of not quarantining when recommended to quarantine due to a positive test 
    \item Quarantine dropout (household) $\in [0.02, 0.05]$: likelihood of not quarantining when recommended to quarantine because a household member got a positive test 
    \item All-levels dropout $\in [0.01, 0.05]$: likelihood of not following app-recommended behavior and instead exhibit pre-pandemic behavior 
\end{enumerate}

\subsection*{App adoption}
COVIsim \cite{gupta2020covisim} models app adoption proportional to smartphone usage statistics. 

\begin{table}[h]
\begin{center}
\begin{tabular}{c|c}
\% of population with app & Uptake required to get that \%  \\
\hline
~1 & ~1.50 \\
30 & 42.15 \\
40 & 56.18 \\
60 & 84.15 \\
70 & 98.31 
\end{tabular}
\end{center}
\caption{
\textbf{Adoption Rate vs Uptake}: The left column show the total percentage of the population with the app, while the right column shows the proportion of \textit{smartphone users} with the app. }
\end{table}

\subsection*{Training time} 
Our ML experiments use approximately 250 days training time on GPUs while simulations required approximately 41 days of CPU time. All CPU time was run on compute using renewable resources.

\end{document}